\documentclass[default,iicol]{sn-jnl}

\jyear{2022}%

\theoremstyle{thmstyleone}%
\theoremstyle{thmstyletwo}%

\theoremstyle{thmstylethree}%

\usepackage{subfig}
\usepackage{acronym}
\usepackage{xspace}
\usepackage{siunitx}
\usepackage{amssymb}
\usepackage{booktabs}
\usepackage{colortbl}
\usepackage{multirow}
\usepackage{rotating}
\usepackage{xcolor}
\usepackage[normalem]{ulem}
\usepackage{array, makecell} %
\makeatletter
\DeclareRobustCommand\onedot{\futurelet\@let@token\@onedot}
\def\@onedot{\ifx\@let@token.\else.\null\fi\xspace}

\def\eg{\emph{e.g}\onedot} 
\def\ie{\emph{i.e}\onedot} 
\def\cf{\emph{c.f}\onedot} 
\def\etc{\emph{etc}\onedot} 
\def\wrt{w.r.t\onedot} 

\acrodef{imu}[IMU]{inertial measurement unit}
\acrodef{fpga}[FPGA]{field-programmable gate array}
\acrodef{slam}[SLAM]{simultaneous localization and mapping}
\acrodef{vio}[VIO]{visual-inertial odometry}
\acrodef{vo}[VO]{visual odometry}
\acrodef{6dof}[6DoF]{six degrees of freedom}
\acrodef{rpe}[RPE]{relative pose error}
\acrodef{rms}[RMS]{root mean square}
\acrodef{sfm}[SfM]{structure from motion}
\acrodef{rtk}[RTK]{real-time kinematic}
\acrodef{gnss}[GNSS]{global navigation satellite system}
\acrodef{ins}[INS]{inertial navigation system}
\acrodef{ate}[ATE]{absolute trajectory error}
\acrodef{lidar}[LiDAR]{light detection and ranging}
\acrodef{ba}[BA]{bundle adjustment}
\acrodef{ransac}[RANSAC]{random sample consensus}
\acrodef{sift}[SIFT]{scale-invariant feature transform}
\acrodef{nn}[NN]{nearest neighbor}
\acrodef{pnp}[PnP]{Perspective-n-Point}

\acused{gnss}
\acused{rtk}
\acused{lidar}
\acused{fpga}

\usepackage{color}

\newcommand{\ra}[1]{\renewcommand{\arraystretch}{#1}}

\newlength{\tabcolsepdefault}
\definecolor{GOLD}{HTML}{FFD700}
\definecolor{SILVER}{HTML}{C0C0C0}
\definecolor{skyblue}{rgb}{0.53, 0.81, 0.92}

\newcommand*{\transpose}{^{\top}}

\usepackage{etoolbox}
\makeatletter
\patchcmd{\ps@headings}
{\hbox to \hsize{\hfill Springer Nature 2021 \LaTeX\ template\hfill}}
{\hbox to \hsize{}}
{}
{}
\patchcmd{\ps@headings}
{\hbox to \hsize{\hfill Springer Nature 2021 \LaTeX\ template\hfill}}
{\hbox to \hsize{}}
{}
{}
\patchcmd{\ps@titlepage}
{\hbox to \hsize{\hfill Springer Nature 2021 \LaTeX\ template\hfill}}
{\hbox to \hsize{}}
{}
{}
\makeatother

\begin{document}

\title[Wenzel et al.]{4Seasons: Benchmarking Visual SLAM and Long-Term Localization for Autonomous Driving in Challenging Conditions}


\author*[1]{\fnm{Patrick} \sur{Wenzel}}\email{patrick.wenzel@tum.de}

\author[2]{\fnm{Nan} \sur{Yang}}
\equalcont{Work done at Technical University of Munich.}

\author[3]{\fnm{Rui} \sur{Wang}}
\equalcont{Work done at Technical University of Munich.}

\author[4]{\fnm{Niclas} \sur{Zeller}}
\equalcont{Work done at Technical University of Munich.}

\author[1]{\fnm{Daniel} \sur{Cremers}}

\affil[1]{Department of Computer Science, Technical University of Munich, Germany}

\affil[2]{Reality Labs at Meta, Redmond, United States}

\affil[3]{Microsoft Mixed Reality \& AI Lab, Zurich, Switzerland}

\affil[4]{Karlsruhe University of Applied Sciences, Karlsruhe, Germany}


\abstract{In this paper, we present a novel visual SLAM and long-term localization benchmark for autonomous driving in challenging conditions based on the large-scale 4Seasons dataset. The proposed benchmark provides drastic appearance variations caused by seasonal changes and diverse weather and illumination conditions. While significant progress has been made in advancing visual SLAM on small-scale datasets with similar conditions, there is still a lack of unified benchmarks representative of real-world scenarios for autonomous driving. We introduce a new unified benchmark for jointly evaluating visual odometry, global place recognition, and map-based visual localization performance which is crucial to successfully enable autonomous driving in any condition. The data has been collected for more than one year, resulting in more than \SI{300}{\km} of recordings in nine different environments ranging from a multi-level parking garage to urban (including tunnels) to countryside and highway. We provide globally consistent reference poses with up to centimeter-level accuracy obtained from the fusion of direct stereo-inertial odometry with RTK GNSS. We evaluate the performance of several state-of-the-art visual odometry and visual localization baseline approaches on the benchmark and analyze their properties. The experimental results provide new insights into current approaches and show promising potential for future research. Our benchmark and evaluation protocols will be available at~\url{https://go.vision.in.tum.de/4seasons}.}


\maketitle

\section{Introduction}\label{sec:introduction}

During the last decade, research on \ac{vo} and \ac{slam} has made tremendous strides~\citep{newcombe2011dtam,engel2014lsd,mur2015orb,engel2017direct}, particularly in the context of autonomous driving~\citep{engel2015stereolsd,wang2017stereoDSO,yang2018dvso,mur2017orb2}. One reason for this progress has been the publication of large-scale datasets tailored for benchmarking these methods~\citep{Cordts2016city,Geiger2013IJRR,Caesar2019nuscenes}. Nonetheless, existing algorithms have significant limitations. Most approaches are tailored to work well on small-scale datasets which exhibit limited challenging conditions. 

Therefore, the next logical step towards progressing research in the direction of visual \ac{slam} is to make it robust under dynamically changing and challenging conditions. This includes \ac{vo}, \eg at night or rain, as well as long-term place recognition and localization against a pre-built map. In this regard, the advent of deep learning has exhibited itself to be a promising potential in complementing the performance of visual \ac{slam}~\citep{Dusmanu2019CVPR,jung2019corl,gn-net-2020,jaramillo2017direct}. Therefore, it has become all the more important to have datasets that are commensurate with handling the challenges of any real-world environment while also being capable of discerning the performance of state-of-the-art approaches. 

\begin{figure}[t]
\centering
\includegraphics[width=0.9\linewidth]{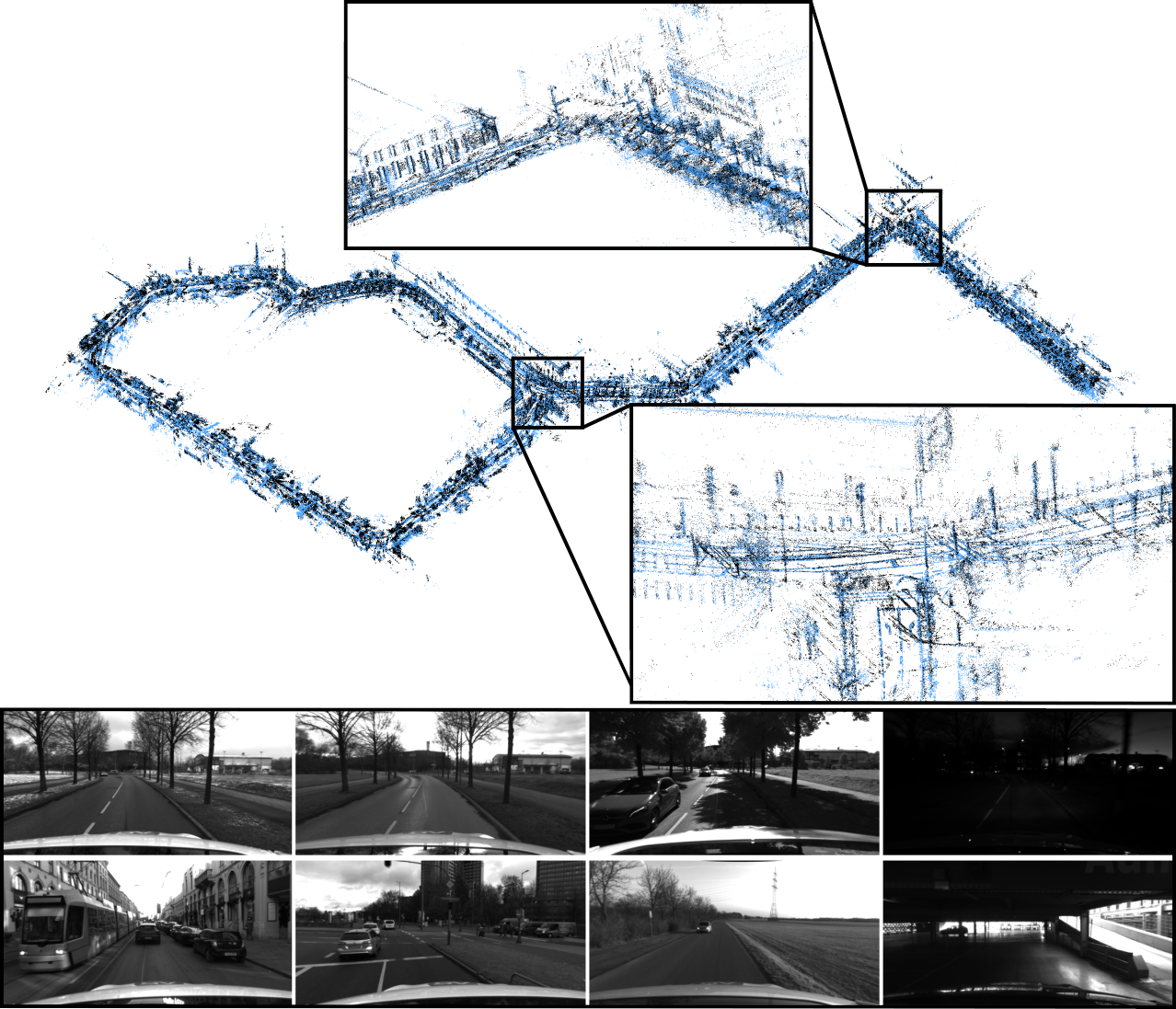}
\caption{\textbf{4Seasons benchmark dataset overview.} Top: overlaid maps recorded at different times and environmental conditions. The 3D points from the reference map (black) align well with the 3D points from the query map (blue), indicating that the reference poses are accurate. Bottom: sample images demonstrating the diversity of our benchmark. The first row shows a collection from the same scene across different weather and lighting conditions: snowy, cloudy, sunny, and night. The second row depicts the variety of scenarios within the benchmark: inner city, suburban, countryside, and a parking garage.}
\label{fig:teaser}
\end{figure}

To accommodate this demand, we present a cross-season and multi-weather benchmark, particularly focusing on visual \ac{slam} and long-term localization for autonomous driving. This benchmark is based on the versatile large-scale 4Seasons dataset~\citep{wenzel2020fourseasons}. To the best of our knowledge, we provide the first large-scale cross-season benchmark dataset comprising stereo images, corresponding high frame-rate \ac{imu}, and accurate \ac{rtk} \ac{gnss} measurements to evaluate sequential localization methods. By traversing the same route under different conditions and over a long-term time horizon, we capture variety in illumination and weather, as well as in the appearance of the scenes. For each scenario, we provide multiple traversals exhibiting different environmental conditions, as described in Table~\ref{tab:4seasons_dataset_with_GT}. The recordings show vastly different variations in the scene geometry including dynamic objects, roadworks, construction sites, and seasonal changes. To acquire accurate reference poses of large-scale scenes, we use a custom stereo-inertial sensor together with a \ac{rtk} \ac{gnss} system to obtain up to centimeter-accurate poses. Figure~\ref{fig:teaser} visualizes two overlaid 3D reconstructions of the same scene recorded at different times. Moreover, the figure depicts sample images of the dataset used to evaluate \ac{6dof} localization against a prior map using query images taken from a variety of challenging conditions. We provide reference poses for a subset of the recordings, and withhold the remaining for an online evaluation benchmark suite. We design a benchmark to measure the impact of long-term environmental changes on the performance of visual \ac{slam} and localization for autonomous driving. 

The main contributions of this paper are the extensive benchmark suite for evaluating the long-term visual localization problem for autonomous driving, the evaluation of state-of-the-art baseline \ac{slam} and visual localization algorithms, and the interpretation of the results.

This work extends our paper published at GCPR 2020~\citep{wenzel2020fourseasons} through the following additional contributions:

\begin{itemize}
\item We propose a large-scale cross-season and multi-weather benchmark suite for long-term visual \ac{slam} in automotive applications. It allows the joint evaluation of \acl{vo}, global place recognition, and map-based visual localization approaches.
\item We release plenty of additional sequences covering nine different types of environments, ranging from a multi-level parking garage to urban (including tunnels) to countryside and highway.
\item We provide an extensive evaluation of state-of-the-art baseline approaches for visual \ac{slam} and visual localization on the presented benchmark.
\end{itemize}

To foster research, our benchmark and evaluation protocols will be available at~\url{https://go.vision.in.tum.de/4seasons}.
\section{Related Work}\label{sec:related_work}
There exists a variety of benchmarks and datasets focusing on \ac{vo} and \ac{slam} for autonomous driving. Here, we divide these datasets into the ones which focus only on \ac{vo} as well as those covering different weather conditions and therefore aiming towards long-term \ac{slam}.

\subsection{Visual Odometry Datasets \& Benchmarks}
The most popular benchmark for autonomous driving probably is KITTI~\citep{Geiger2013IJRR}. This multi-sensor dataset covers a wide range of tasks including not only \ac{vo}, but also 3D object detection and tracking, scene flow estimation as well as semantic scene understanding. The dataset contains diverse scenarios ranging from urban to countryside to highway. Nevertheless, all scenarios are only recorded once and under similar weather conditions. Ground truth is obtained based on a high-end \ac{ins}.

Another dataset containing \ac{lidar}, \ac{imu}, and image data at a large scale is the Málaga Urban dataset~\citep{Blanco2014malaga}. However, in contrast to KITTI, no accurate \ac{6dof} ground truth is provided, and therefore it does not allow for an appropriate quantitative evaluation. Moreover, only a few places are visited multiple times.

Other popular datasets for the evaluation of \ac{vo} and \ac{vio} algorithms that are not related to autonomous driving include~\citep{sturm12iros} (handheld RGB-D),~\citep{burri2016euroc} (UAV stereo-inertial),~\citep{engel2016monodataset} (handheld mono), and~\citep{schubert2018vidataset} (handheld stereo-inertial).

\subsection{Long-Term SLAM Datasets \& Benchmarks}
More related to our work are datasets containing multiple traversals of the same environment over a long period. Concerning \ac{slam} for autonomous driving, the Oxford RobotCar Dataset~\citep{maddern20171} represents a kind of pioneer work. This dataset consists of large-scale sequences recorded multiple times in the same environment for one year. Hence, it covers large variations in the appearance and structure of the scene. However, the diversity of the scenarios is only limited to an urban environment. Also, the ground truth provided for the dataset is not accurate up to centimeter-level~\citep{maddern20171,spencer2020same}. Other existing datasets are lacking sequential structure~\citep{kenk2020dawn}, only provide a certain adverse condition~\citep{pitropov2021canadian}, or focus on AR scenarios~\citep{sarlin2022lamar}. 

The work by~\citep{SattlerCVPR2018} proposes three complementary benchmark datasets based on existing datasets, namely RobotCar Seasons (based on~\citep{maddern20171}), Aachen Day-Night (based on~\citep{sattler2012image}), and CMU Seasons (based on~\citep{5940504}) that have been used for benchmarking visual localization approaches. The ground truth of the RobotCar Seasons~\citep{SattlerCVPR2018} dataset is obtained based on \ac{sfm} and \ac{lidar} point cloud alignment. However, due to inaccurate \ac{gnss} measurements~\citep{maddern20171}, a globally consistent ground truth up to centimeter-level accuracy can not be guaranteed. Furthermore, this dataset only provides one reference traversal in the overcast condition. In contrast, we provide globally consistent reference models for all training traversals covering a wide variety of conditions. Hence, every traversal can be used as a reference model that allows further research on, \eg analyzing suitable reference-query pairs for long-term localization and mapping. 

Global place recognition datasets such as Pittsburgh~\citep{torii2013visual}, Tokyo 24/7~\citep{torii201524}, and Mapillary Street-Level Sequences~\citep{warburg2020mapillary} provide only coarse-scale location information. Other related localization datasets include 12-Scenes~\citep{valentin2016learning}, InLoc~\citep{taira2018inloc}, Cambridge Landmarks~\citep{kendall2015posenet}, and CrowdDriven~\citep{jafarzadeh2021crowddriven}.

\subsection{Other Datasets}
Examples of further multipurpose autonomous driving datasets that also can be used for \ac{vo} are~\citep{Cordts2016city,Wang2017toronto,Huang2018apollo,Caesar2019nuscenes}.

As stated in Section~\ref{sec:introduction}, our proposed benchmark dataset differentiates from previous related work in terms of being both large-scale (similar to~\citep{Geiger2013IJRR}) and having high variations in appearance and conditions (similar to~\citep{maddern20171}). Furthermore, accurate reference poses based on the fusion of direct stereo \ac{vio} and \ac{rtk} \ac{gnss} are provided. To the best of our knowledge, we are the first to introduce a public, modular benchmark for evaluating visual \ac{slam}, global place recognition, and map-based visual localization approaches under challenging conditions for autonomous driving.

\section{System Overview}\label{sec:overview}

\begin{figure*}[t]
\centering
\subfloat[Test vehicle.]{\includegraphics[width=0.45\linewidth]{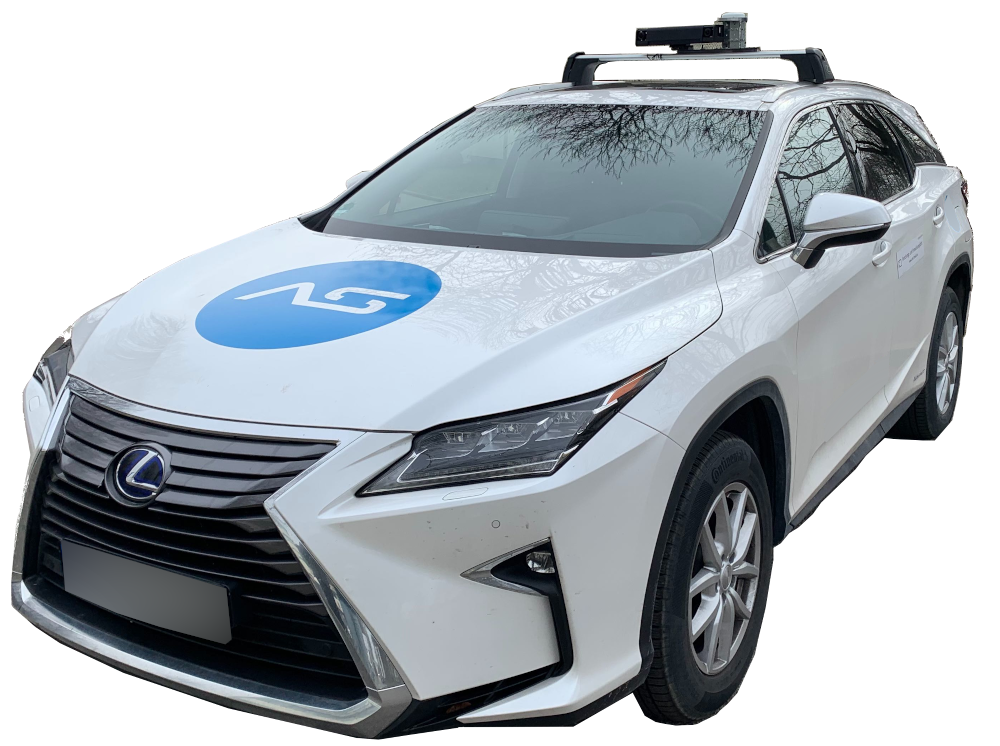}}\qquad
\subfloat[Sensor system.]{\includegraphics[width=0.45\linewidth]{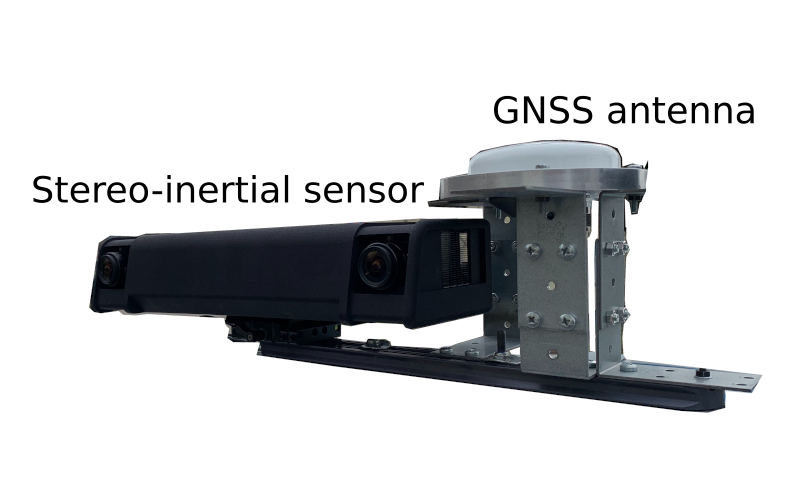}}
\caption{\textbf{Recording setup.} Test vehicle and sensor system used for dataset recording. The sensor system consists of a custom stereo-inertial sensor with a stereo baseline of \SI{30}{\cm} and a high-end \ac{rtk} \ac{gnss} receiver from Septentrio.}
\label{fig:test_vehicle}
\end{figure*}

This section presents the sensor setup which is used for data recording (Section~\ref{sec:sensor_setup}). Furthermore, we describe the calibration of the entire sensor suite (Section~\ref{sec:calibration}) as well as our approach to obtain up to centimeter-accurate global \ac{6dof} reference poses (Section~\ref{sec:ground_truth}).

\subsection{Sensor Setup}\label{sec:sensor_setup}
The hardware setup consists of a custom stereo-inertial sensor for \ac{6dof} pose estimation, as well as a high-end \ac{rtk} \ac{gnss} receiver for global positioning and global pose refinement. Figure~\ref{fig:test_vehicle} shows our test vehicle equipped with the sensor system used for data acquisition.

\subsubsection{Stereo-Inertial Sensor}
The core of the sensor system is our custom stereo-inertial sensor. This sensor consists of a pair of monochrome industrial-grade global shutter cameras (Basler acA2040-35gm) and lenses with a fixed focal length of $f=\SI{3.5}{\mm}$ (Stemmer Imaging CVO GMTHR23514MCN). The cameras are mounted on a highly-rigid aluminum rail with a stereo baseline of \SI{30}{\cm}. On the same rail, a Precision MEMS \ac{imu} (Analog Devices ADIS16465) is mounted. The cameras and the \ac{imu} are triggered over an external clock generated by an \ac{fpga}. Here, the trigger accounts for exposure compensations, meaning that the time between the centers of the exposure interval for two consecutive images is always kept constant (1/[frame rate]) independent of the exposure time itself.

Furthermore, based on the \ac{fpga}, the \ac{imu} is properly synchronized with the cameras. In the dataset, we record stereo sequences with a frame rate of \SI{30}{fps}. We perform pixel binning with a factor of two and crop the image to a resolution of $800 \times 400$. This results in a field of view of approximately \ang{77} horizontally and \ang{43} vertically. The \ac{imu} is recorded at a frequency of \SI{2000}{\Hz}. During recording, we guarantee an equal exposure time for the left and the right image of each stereo pair as well as a smooth exposure transition in highly dynamic lighting conditions, as it is favorable to visual \ac{slam}. We provide those exposure times for each frame. 

\subsubsection{GNSS Receiver}
For global positioning and to compensate drift in the \ac{vio} system, we utilize an \ac{rtk} \ac{gnss} receiver (mosaic-X5) from Septentrio in combination with an Antcom Active G8 \ac{gnss} antenna. The \ac{gnss} receiver provides a horizontal position accuracy of up to \SI{6}{\mm} by utilizing \ac{rtk} correction signals. While the high-end \ac{gnss} receiver is used for accurate positioning, we use a second receiver connected to the time-synchronization \ac{fpga} to obtain \ac{gnss} timestamps for the sensors.

\subsection{Calibration}\label{sec:calibration}

\subsubsection{Aperture and Focus Adjustment}
The lenses used in the stereo system have both adjustable aperture and focus. Therefore, before performing the geometric calibration of all sensors, we manually adjust both cameras for a matching average brightness and a minimum focus blur~\citep{Hu2006ICIP}, across a structured planar target in \SI{10}{\m} distance.

\subsubsection{Stereo Camera and IMU}
For the intrinsic and extrinsic calibration of the stereo cameras, as well as the extrinsic calibration and time-synchronization of the \ac{imu}, we use Kalibr\footnote{\url{https://github.com/ethz-asl/kalibr}}~\citep{rehder2016extending}. The stereo cameras are modeled using the Kannala-Brandt model~\citep{Kannala2006}, a generic camera model consisting of a total of eight parameters. We validated the calibration accuracy of each recording by performing a feature-based epipolar-line consistency check.

\subsubsection{GNSS Antenna}
Since the \ac{gnss} antenna does not have any orientation but has an isotropic reception pattern, only the 3D translation vector between one of the cameras and the antenna within the camera frame has to be known. This vector was measured manually for our sensor setup.

\subsection{Ground Truth Generation}\label{sec:ground_truth}
Reference poses (\ie ground truth) for \ac{vo} and \ac{slam} should provide high accuracy in both local relative \ac{6dof} transformations and global positioning. To fulfill the first requirement, we extend the state-of-the-art stereo direct sparse \ac{vo}~\citep{wang2017stereoDSO} by integrating \ac{imu} measurements~\citep{von2018direct}, achieving a stereo-inertial \ac{slam} system offering average tracking drift around \SI{0.6}{\percent} of the traveled distance.

To fulfill the second requirement, the poses estimated by our stereo-inertial system are fused with the \ac{rtk} \ac{gnss} measurements using a global pose graph. We first estimate a $\mathrm{Sim}(3)$ transformation to globally align the camera positions in the \ac{vio} coordinate system to those in the \ac{gnss} coordinate system using the Kabsch–Umeyama algorithm~\citep{umeyama1991least}. A transformation in $\mathrm{Sim}(3)$ is estimated instead of in $\mathrm{SE}(3)$ to account for the global scale drift in the \ac{vio} system. Denoting the Lie-algebra of $\mathrm{SE}(3)$ as $\mathfrak{se}(3)$, each aligned camera pose $\boldsymbol{\xi}^{\text{VIO}}_{wi} \in \mathfrak{se}(3)$ is added to the pose graph as a $\mathfrak{se}(3)$ node, where $\boldsymbol{\xi}_{wi}$ defines a transformation from the $i$-th camera coordinate system to the world coordinate system. The camera connections from the \ac{vio} sliding window (one connection corresponds to two cameras co-observing a part of the scene) are added as $\mathfrak{se}(3)-\mathfrak{se}(3)$ edges, with the relative poses $\boldsymbol{\xi}^{\text{VIO}}_{ji}$ as the measurements. If a camera pose has a valid corresponding \ac{gnss} pose, that is, the \ac{gnss} pose is available and the observed standard deviation of the position is smaller than a predefined threshold, the \ac{gnss} pose $\mathbf{t}_{i} \in \mathbb{R}^3$ is added to the pose graph as a fixed $\mathbb{R}^3$ node and an $\mathfrak{se}(3)-\mathbb{R}^3$ edge is added. The energy function defined for the pose graph optimization is thus defined as:
\begin{equation}
\begin{aligned}
    & E(\boldsymbol{\xi}_{wi},\dots,\boldsymbol{\xi}_{wn}) = \\ 
    & \sum_{\boldsymbol{\xi}^{\text{VIO}}_{ji} \in \mathbb{\varepsilon}}(\boldsymbol{\xi}^{\text{VIO}}_{ji} \circ  \boldsymbol{\xi}^{-1}_{wi} \circ \boldsymbol{\xi}_{wj})\transpose \mathbf{\Sigma}^{-1}_{ji} (\boldsymbol{\xi}^{\text{VIO}}_{ji} \circ  \boldsymbol{\xi}^{-1}_{wi} \circ \boldsymbol{\xi}_{wj}) +\\
    & \omega \sum_{\mathbf{t}_{i} \in \mathbb{\nu}} (\mathbf{t}_{i} - (\boldsymbol{\xi}_{wi} \circ \boldsymbol{\xi}_{cg})^{[\mathbf{t}]})\transpose \mathbf{\Sigma}^{-1}_{i} (\mathbf{t}_{i} - (\boldsymbol{\xi}_{wi} \circ \boldsymbol{\xi}_{cg})^{[\mathbf{t}]}),
\end{aligned}
\label{eq:pgo_energy_function}
\end{equation}
where $\mathbb{\varepsilon}$ is the set of \ac{vio} camera connections, $\mathbb{\nu}$ is the set of valid \ac{rtk} \ac{gnss} poses. $\mathbf{\Sigma}_{ji} \in \mathbb{R}^{6\times6}$ and $\mathbf{\Sigma}_{i} \in \mathbb{R}^{3\times3}$ are the covariance matrices from the \ac{vio} and \ac{gnss} systems. $\boldsymbol{\xi}_{cg}$ denotes the extrinsic calibration between the camera and the \ac{gnss} antenna. A scale term $\omega$ is added to balance the two different domains. The $\circ$-operator defines the concatenation of poses defined as $\mathfrak{se}(3)$ and therefore is defined as follows:
\begin{align}
\boldsymbol{\xi}_{i} \circ  \boldsymbol{\xi}_{j} := \log(\exp(\boldsymbol{\xi}_{i})\cdot \exp(\boldsymbol{\xi}_{j})),
\end{align}
where $\log(\cdot)$ defines the logarithm and $\exp(\cdot)$ the exponential map of the $\mathrm{SE}(3)$ Lie-algebra. $\boldsymbol{\xi}^{[\mathbf{t}]}$ denotes the translation part in $\mathfrak{se}(3)$. The energy function is optimized using the Levenberg–Marquardt algorithm in~\citep{kummerle2011g}.

One crucial aspect of the dataset is that the reference poses that we provide are accurate enough, even though some recorded sequences contain challenging conditions in partially \ac{gnss}-denied environments. Although the stereo-inertial sensor system has an average drift around \SI{0.6}{\percent}, this cannot be guaranteed for all cases. Hence, for the reference poses in our dataset, we report whether a pose can be considered to be reliable by measuring the distance to the corresponding \ac{rtk} \ac{gnss} measurement. For all poses, without corresponding \ac{rtk} \ac{gnss} measurement we do not guarantee a certain accuracy. Nevertheless, due to the highly accurate stereo-inertial odometry system, these poses can be considered accurate in most cases, even in environments without \ac{gnss}, \eg tunnels, or areas with tall buildings. We provide details about the pose accuracy in Section~\ref{sec:pose_accuracy}.

\section{Benchmark Setup}\label{sec:benchmark_setup}

To overcome the shortcomings of existing benchmarks and datasets for autonomous driving, as discussed in Section~\ref{sec:related_work}, we define the following requirements for an appropriate benchmark.

\begin{itemize}
    \item \emph{Accuracy:} we provide up to centimeter-accurate \ac{6dof} poses obtained by fusing \ac{vio} measurements with \ac{rtk} \ac{gnss} correction data.
    \item \emph{Large-scale:} we provide large-scale sequences (trajectories longer than \SI{10}{\km}) to allow for extensive evaluations of \ac{slam} and visual localization under challenging conditions.
    \item \emph{Diversity:} besides large-scale, we also provide both short-term and long-term changes within the recorded scenes. This is important to evaluate the generalization capabilities of recent learning-based methods.
    \item \emph{Multitask:} the benchmark can be used to evaluate \acl{vo}, global place recognition, and map-based visual localization under challenging conditions.
\end{itemize}

Based on these properties, we propose a novel large-scale dataset that is used as an extensive benchmark suite for evaluating multitasking challenges related to autonomous driving under changing conditions. The sequences have been collected in the metropolitan area of Munich, Germany. The different scenes are described in the next section.

\begin{table*}[t]
\centering
\ra{1.3}
\caption{\textbf{Statistics of the 4Seasons benchmark.} This table shows the different scenarios and recordings along with the weather condition, seasons, and time of the day from our benchmark. We provide a variety of scenarios and short-term to long-term changes. The recordings in this table are used for the benchmark evaluation. The ground truth (\ac{gnss}/\ac{imu}, point clouds, and reference poses) is withheld. Benchmark type (VO = visual odometry, GPR = global place recognition, MBVL = map-based visual localization) defines the benchmark a sequence is used for. All recordings with ground truth are shown in Table~\ref{tab:4seasons_dataset_with_GT}.}
\label{tab:4seasons_dataset_without_GT}
\resizebox{\textwidth}{!}{
\begin{tabular}{llllllll}
\toprule
\bfseries Scenario & \bfseries Recording & \bfseries \makecell[l]{Weather \\ (cloudy, rainy, snowy, sunny)} & \bfseries \makecell[l]{Season \\ (winter, spring, summer, fall)} & \bfseries \makecell[l]{Daytime \\ (morning, afternoon, evening, night)} & \bfseries Benchmark Type & \bfseries \makecell[l]{Map Accuracy \\ Horizontal RMSE \\ (\ac{gnss}-Ref. Pose)} & \bfseries \makecell[l]{Map Accuracy \\ \% of Accurate Poses}\\
\midrule
office\_loop\_1\_test & 2020-03-03{\_}12-12-32 & cloudy & spring & afternoon & GPR & \SI{12.29}{\cm} & \SI{59.91}{\percent} \\
office\_loop\_2\_test & 2020-03-26{\_}15-03-02 & cloudy/sunny & spring & afternoon & \ac{vo} & \SI{5.14}{\cm} & \SI{90.22}{\percent} \\
office\_loop\_3\_test & 2021-05-10{\_}19-25-54 & cloudy & spring & evening & \ac{vo} + MBVL & \SI{5.78}{\cm} & \SI{92.06}{\percent}\\
\midrule
highway\_1\_test & 2020-10-08{\_}10-19-46 & sunny & fall & morning & \ac{vo} & \SI{8.04}{\cm} & \SI{73.65}{\percent}\\
highway\_2\_test & 2021-02-25{\_}13-11-30 & sunny & winter & afternoon & \ac{vo} & \SI{4.80}{\cm} & \SI{74.31}{\percent}\\
\midrule
neighborhood\_1\_test & 2020-03-26{\_}14-54-05 & cloudy & spring & afternoon & GPR & \SI{2.20}{\cm} & \SI{87.38}{\percent}\\
neighborhood\_2\_test & 2021-05-10{\_}18-26-26 & cloudy & spring & evening & \ac{vo} + MBVL & \SI{1.51}{\cm} & \SI{87.42}{\percent} \\
\midrule
business\_campus\_1\_test & 2021-01-07{\_}13-03-56 & cloudy/snowy & winter & afternoon & \ac{vo} + MBVL & \SI{3.39}{\cm} & \SI{97.36}{\percent}  \\
\midrule
countryside\_1\_test & 2020-03-26{\_}14-30-52 & cloudy & spring & afternoon & GPR & \SI{2.53}{\cm} & \SI{91.75}{\percent} \\
countryside\_2\_test & 2021-01-07{\_}14-03-57 & cloudy/snowy & winter & afternoon & \ac{vo} + MBVL & \SI{2.36}{\cm} & \SI{92.21}{\percent} \\
\midrule
city\_loop\_1\_test & 2020-03-03{\_}12-28-45 & cloudy & spring & afternoon & GPR & \SI{5.36}{\cm} & \SI{83.62}{\percent}\\
city\_loop\_2\_test & 2021-02-25{\_}11-27-40 & sunny & winter & morning & \ac{vo} + MBVL & \SI{3.36}{\cm} & \SI{81.40}{\percent}\\ 
\midrule
old\_town\_1\_test & 2020-10-08{\_}12-11-19 & cloudy & fall & afternoon & GPR & \SI{7.19}{\cm} & \SI{94.26}{\percent}\\
old\_town\_2\_test & 2021-05-10{\_}19-51-14 & cloudy & spring & evening & \ac{vo} & \SI{1.84}{\cm} & \SI{96.04}{\percent} \\ 
old\_town\_3\_test & 2021-05-10{\_}21-18-00 & cloudy & spring & night & \ac{vo} + MBVL & \SI{4.94}{\cm} & \SI{92.07}{\percent}\\
\midrule
maximilianeum\_1\_test & 2021-02-25{\_}12-16-32 & sunny & winter & afternoon & \ac{vo} & \SI{1.90}{\cm} & \SI{80.13}{\percent}\\ 
maximilianeum\_2\_test & 2021-05-10{\_}20-59-00 & cloudy & spring & night & \ac{vo} & \SI{12.46}{\cm} & \SI{76.46}{\percent}\\
\midrule 
parking\_garage\_1\_test & 2020-06-12{\_}10-29-20 & sunny & summer & morning & \ac{vo} + MBVL& \SI{0.75}{\cm} & \SI{35.06}{\percent}\\
parking\_garage\_2\_test & 2021-05-10{\_}19-18-36 & cloudy & spring & evening & GPR & \SI{4.54}{\cm} & \SI{40.75}{\percent}\\
\bottomrule
\end{tabular}
}
\end{table*}

\begin{figure}[t]
    \centering
    \includegraphics[width=0.5\linewidth]{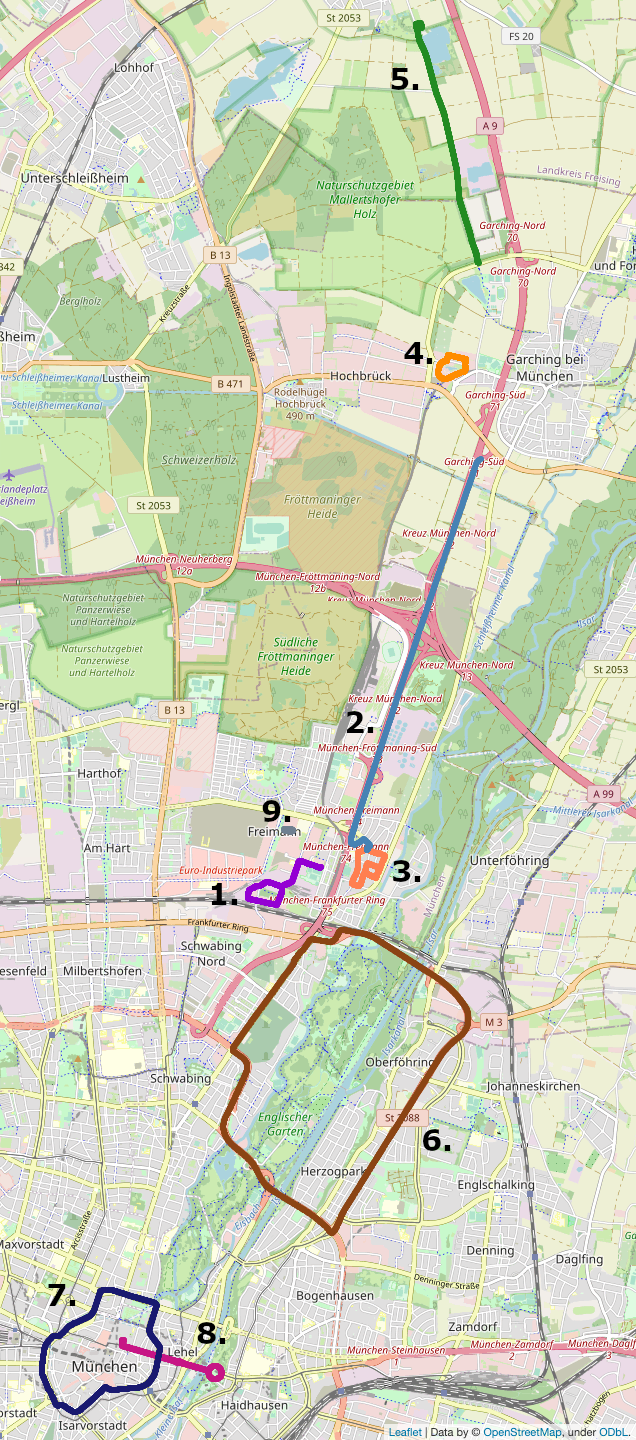}
    \caption{\textbf{Data collection map.} This figure shows the map of the covered area of our benchmark dataset. We provide sequences at a large scale and a huge variety of different environments. A detailed visualization of each scenario's trajectory is shown in Figure~\ref{fig:app_all_scenarios}.}
    \label{fig:map_scenerios}
\end{figure}

\subsection{Scenarios}\label{sec:scenarios}

This section describes the different sequences we have collected for the dataset. The sequences involve different scenarios -- ranging from urban driving to a parking garage and rural areas. We provide complex trajectories, which include partially overlapping routes, and multiple loops within a sequence. For each scenario, we have collected multiple traversals covering a large range of variations in the structure and environmental appearance due to weather, illumination, dynamic objects, and seasonal effects. In total, our benchmark dataset consists of nine different scenarios.

Figure~\ref{fig:map_scenerios} shows the covered area, including highlighted traces. Each scenario is visualized in a separate color. We now describe each scene in more detail.

\begin{enumerate}
    \item \textbf{Office Loop.} A loop around an industrial area of the city.
    \item \textbf{Highway.} A drive along the A9 three-lane highway in the northern part of Munich.
    \item \textbf{Neighborhood.} Traversal through a neighborhood at the outskirts of the city, covering detached houses with gardens and trees in the street.
    \item \textbf{Business Campus.} Several loops around a campus in a business area. 
    \item \textbf{Countryside.} Rural area around agricultural fields that exhibits very homogeneous and repetitive structures.
    \item \textbf{City Loop.} A large-scale loop at a ring road within the city of Munich, including a tunnel.
    \item \textbf{Old Town.} Loop around the urban city center with tall buildings, much traffic, and dynamic objects. 
    \item \textbf{Maximilianeum.} The Maximilianeum is a famous palatial building in Munich which is located at the eastern end of a royal avenue with paving stones and a tram route. 
    \item \textbf{Parking Garage.} A three-level parking garage to benchmark combined indoor and outdoor environments.
\end{enumerate}

\begin{figure*}[t]
    \centering
    \includegraphics[width=0.9\linewidth]{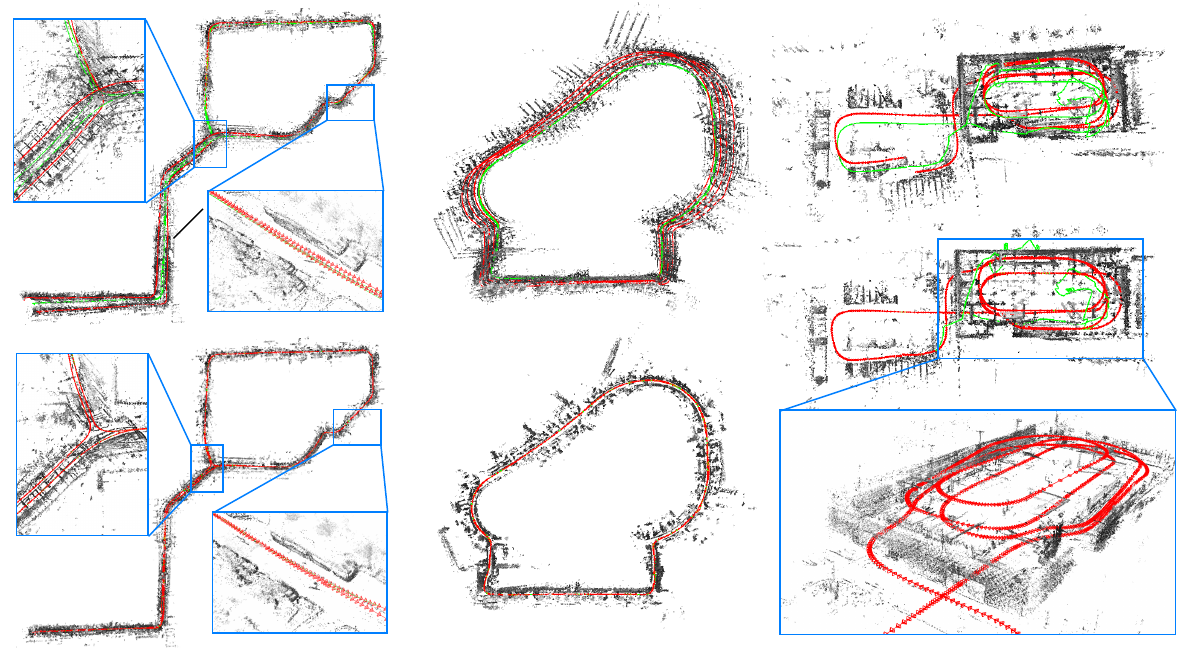}
    \caption{\textbf{3D models of different scenarios contained in the dataset}. The figure shows an office loop around an industrial area (left), multiple loops around a business campus with high buildings (middle), and a stretch recorded in a multi-level parking garage (right). The green lines encode the \ac{gnss} trajectories, and the red lines encode the \ac{vio} trajectories. Top: shows the trajectories before the fusion using pose graph optimization. Bottom: shows the results after the pose graph optimization. Note that after the pose graph optimization, the reference trajectories are well aligned.}
    \label{fig:3d_models}
\end{figure*}

\begin{figure}[t]
    \centering
    \includegraphics[width=0.9\linewidth]{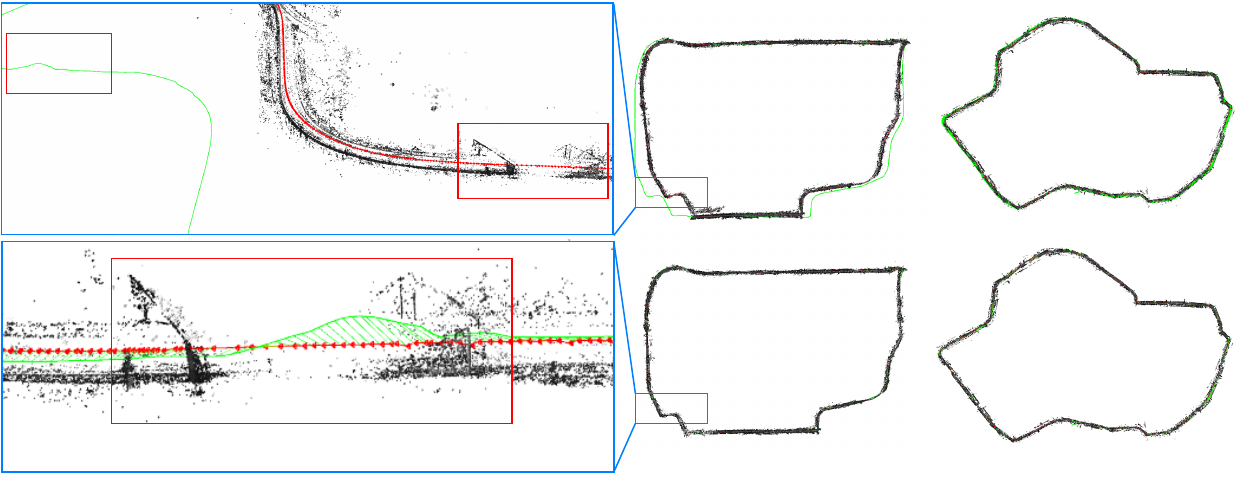}
    \caption{\textbf{Reference poses validation.} This figure shows two additional 3D models of the scenarios collected. Note that these two sequences are quite large (more than \SI{10}{\km} and \SI{6}{\km}, respectively). Top: before the fusion using pose graph optimization. Bottom: results after optimization. The green lines encode the \ac{gnss} trajectories, the red lines show the \ac{vio} trajectories (before fusion) and the fused trajectories (after fusion). The left part of the figure shows a zoomed-in view of a tunnel, where the \ac{gnss} signal becomes very noisy, as highlighted in the red boxes. Besides, due to the large size of the sequence, the accumulated tracking error leads to a significant deviation of the \ac{vio} trajectory from the \ac{gnss} recordings. Our pose graph optimization, by depending globally on \ac{gnss} positions and locally on \ac{vio} relative poses, successfully eliminates global \ac{vio} drifts and local \ac{gnss} positioning flaws.}
    \label{fig:3d_models_supplementary}
\end{figure}

\begin{figure*}%
\renewcommand*\thesubfigure{\arabic{subfigure}} 
\centering
\subfloat[][Office Loop.]{\includegraphics[width=0.3\linewidth]{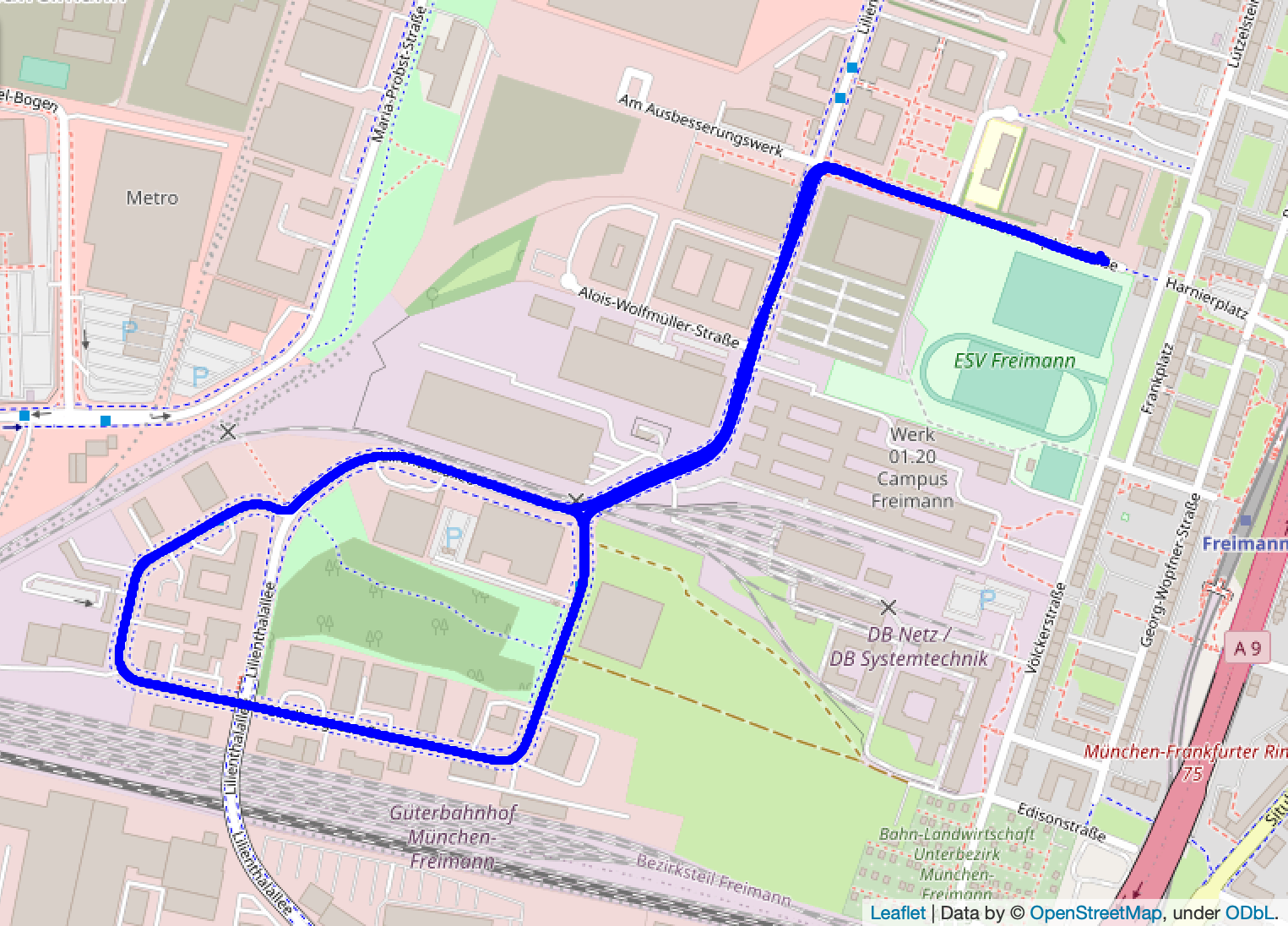}}
\qquad
\subfloat[][Highway.]{\includegraphics[width=0.3\linewidth]{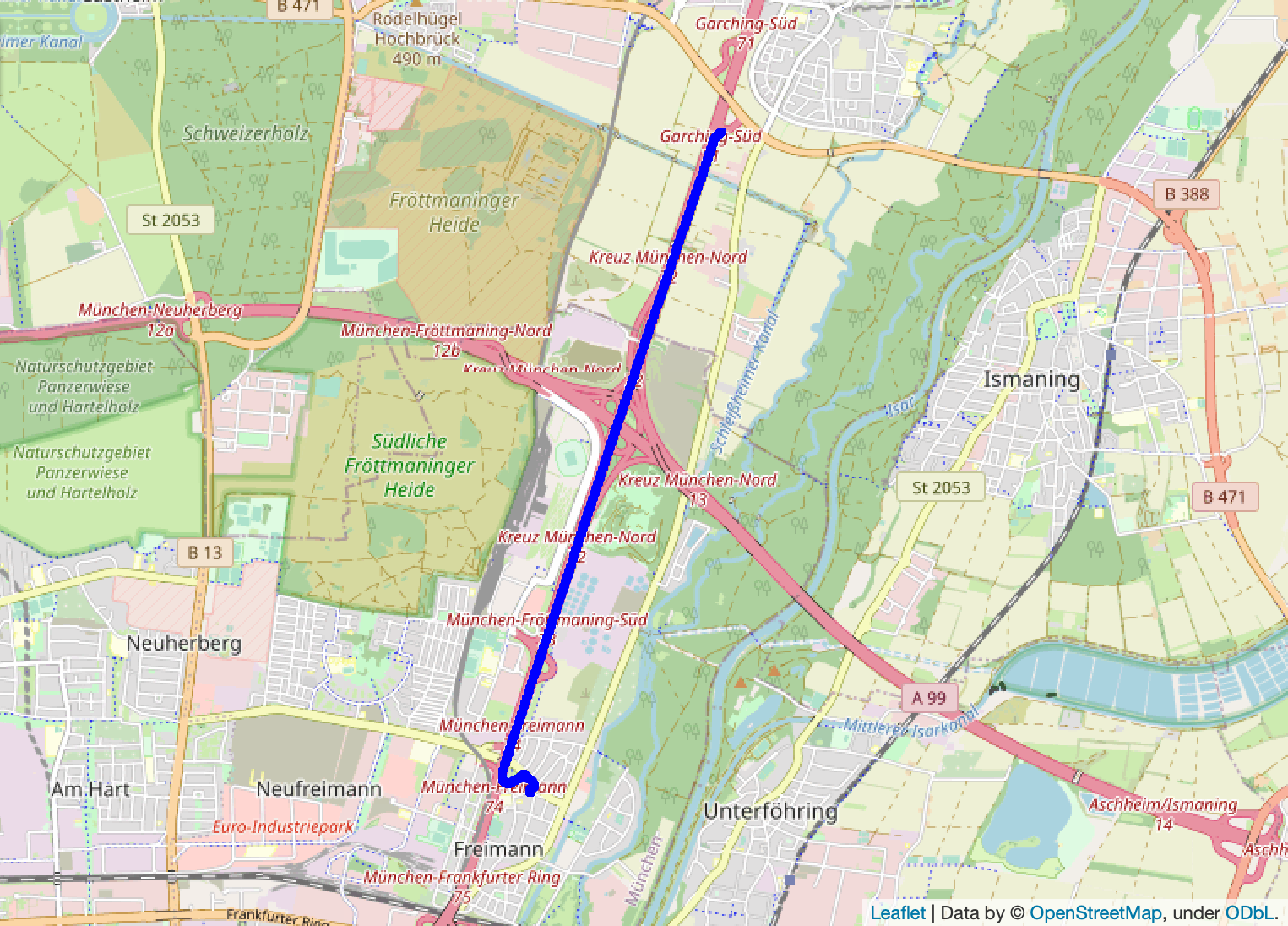}}
\qquad
\subfloat[][Neighborhood.]{\includegraphics[width=0.3\linewidth]{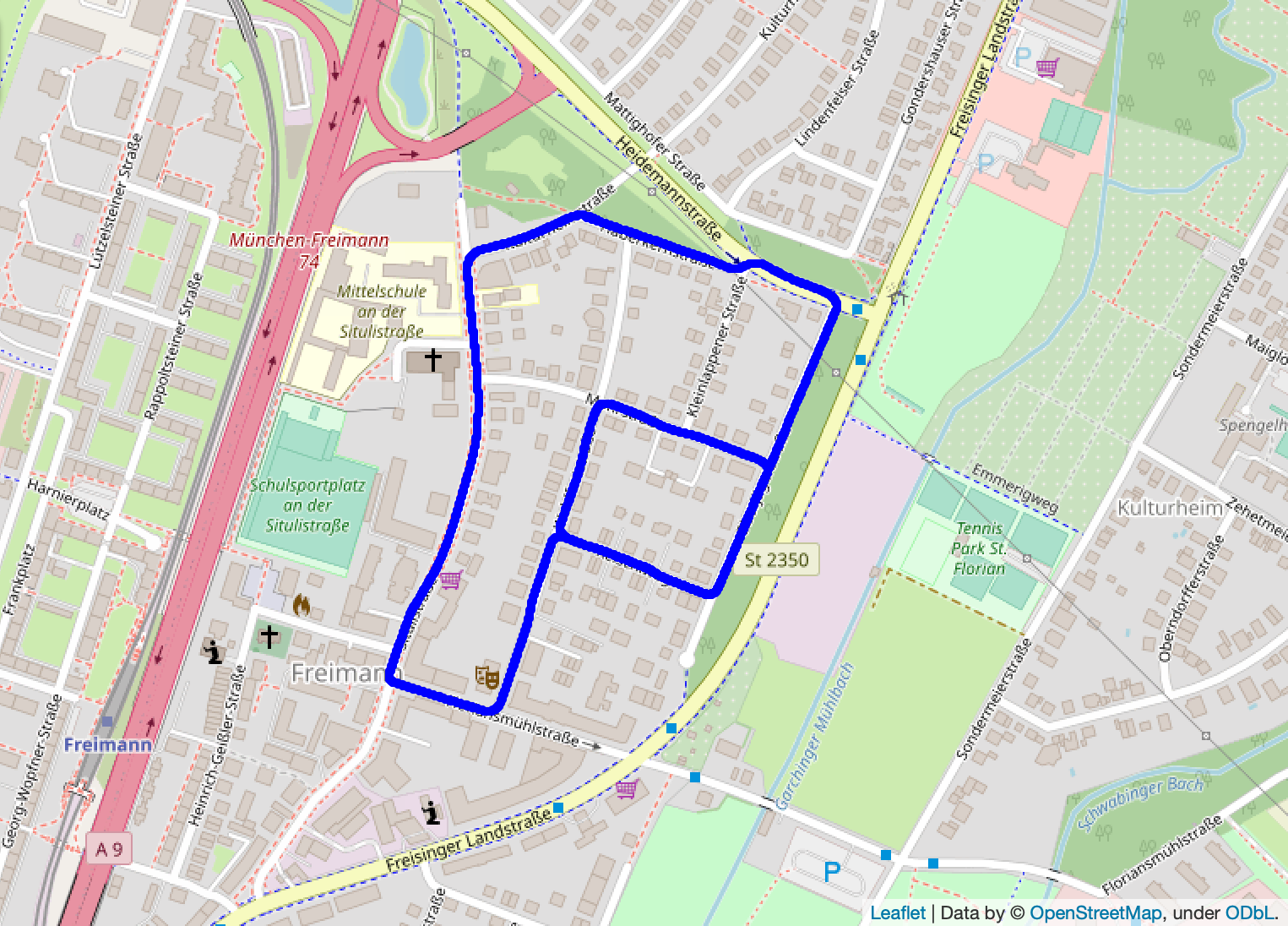}}

\subfloat[][Business Campus.]{\includegraphics[width=0.3\linewidth]{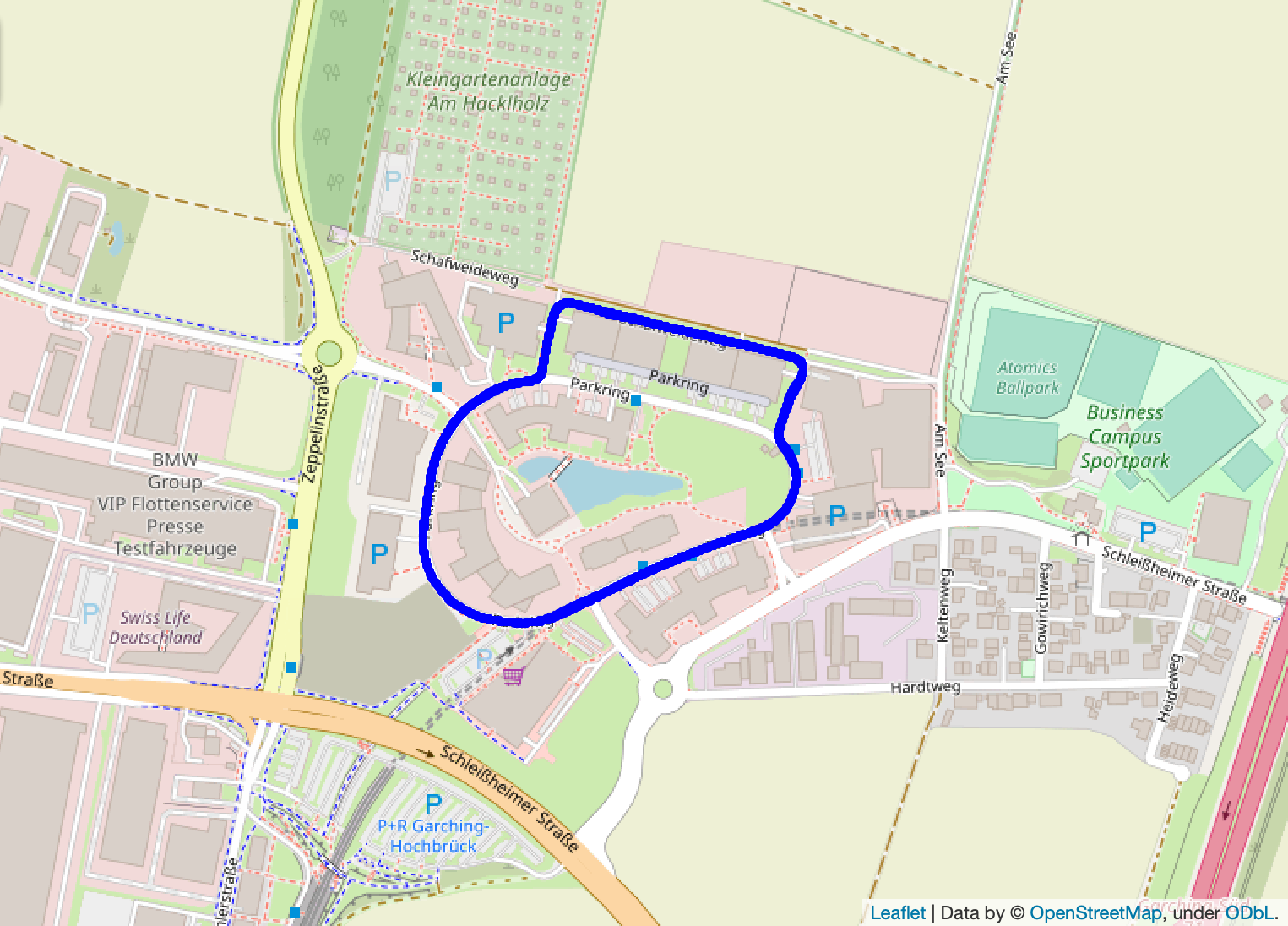}}
\qquad
\subfloat[][Countryside.]{\includegraphics[width=0.3\linewidth]{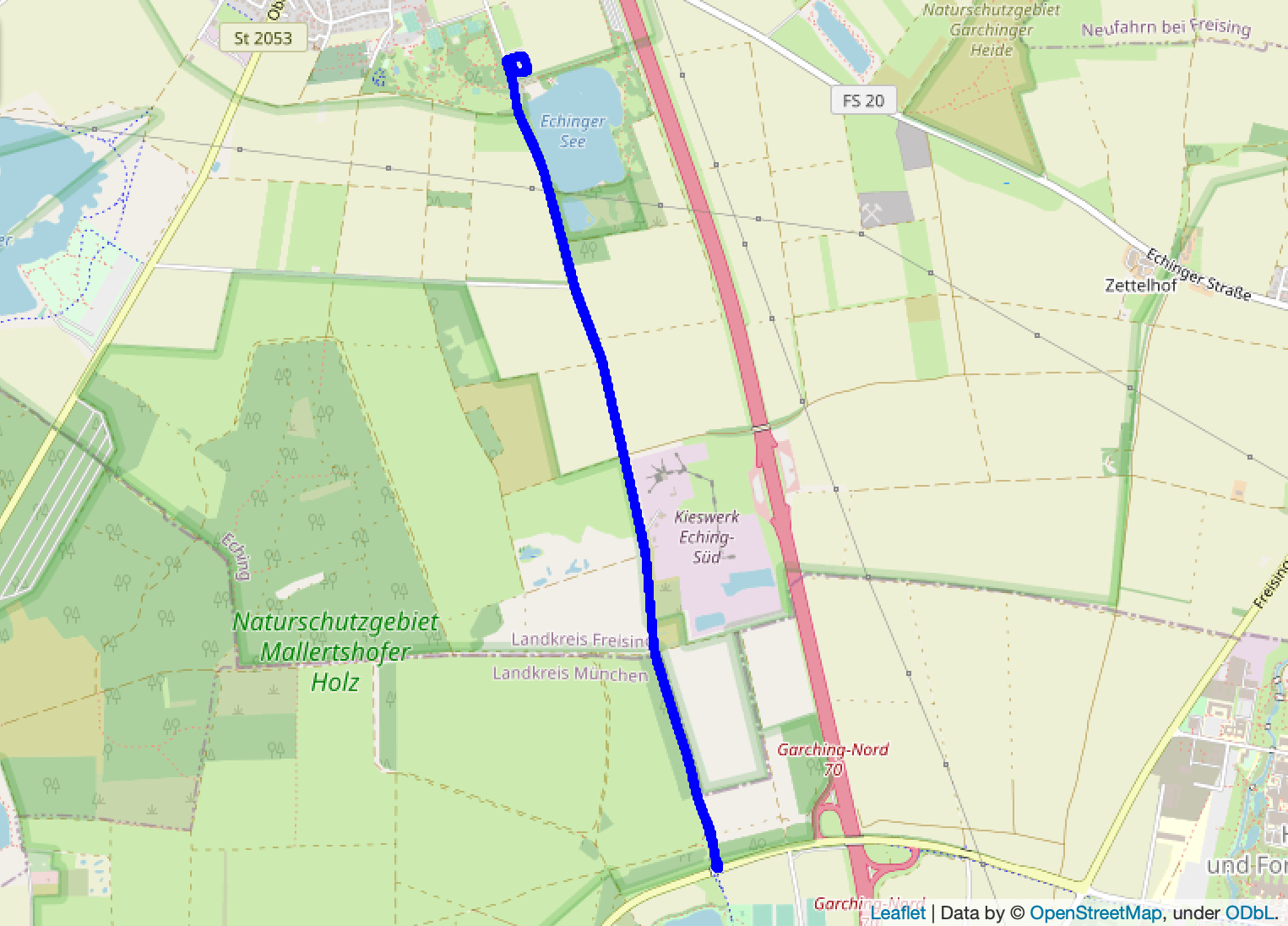}}
\qquad
\subfloat[][City Loop.]{\includegraphics[width=0.3\linewidth]{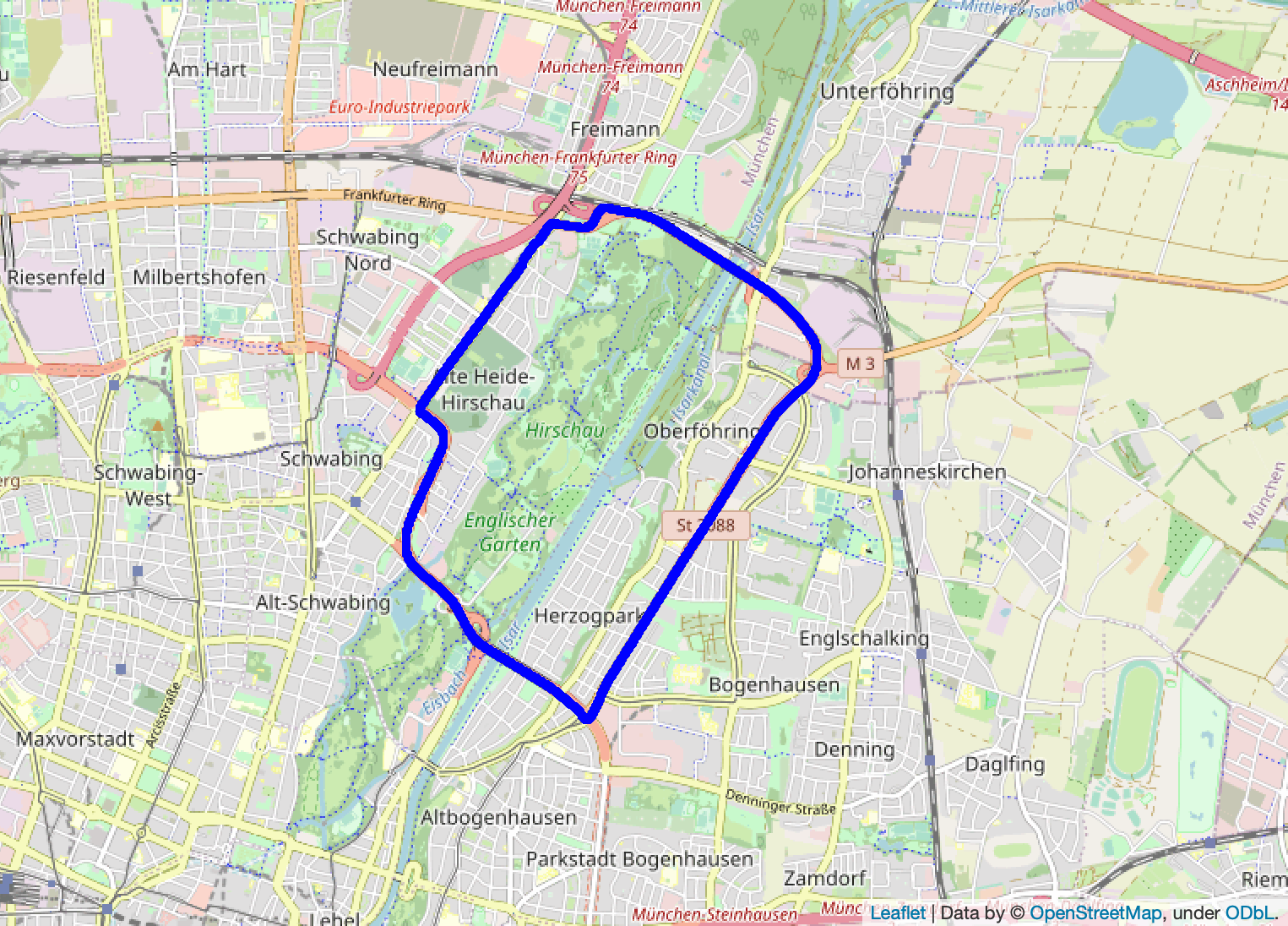}}

\subfloat[][Old Town.]{\includegraphics[width=0.3\linewidth]{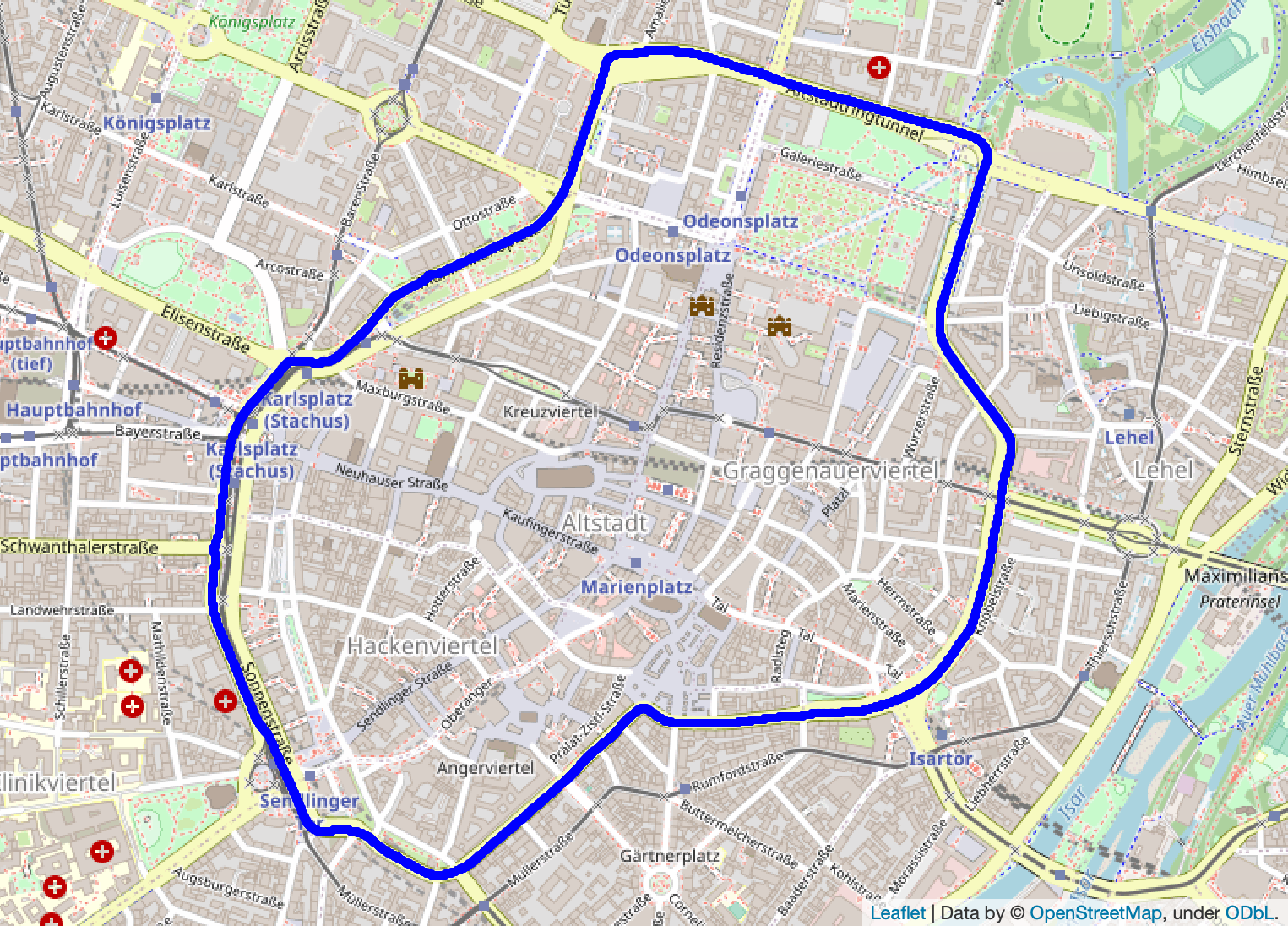}}
\qquad
\subfloat[][Maximilianeum.]{\includegraphics[width=0.3\linewidth]{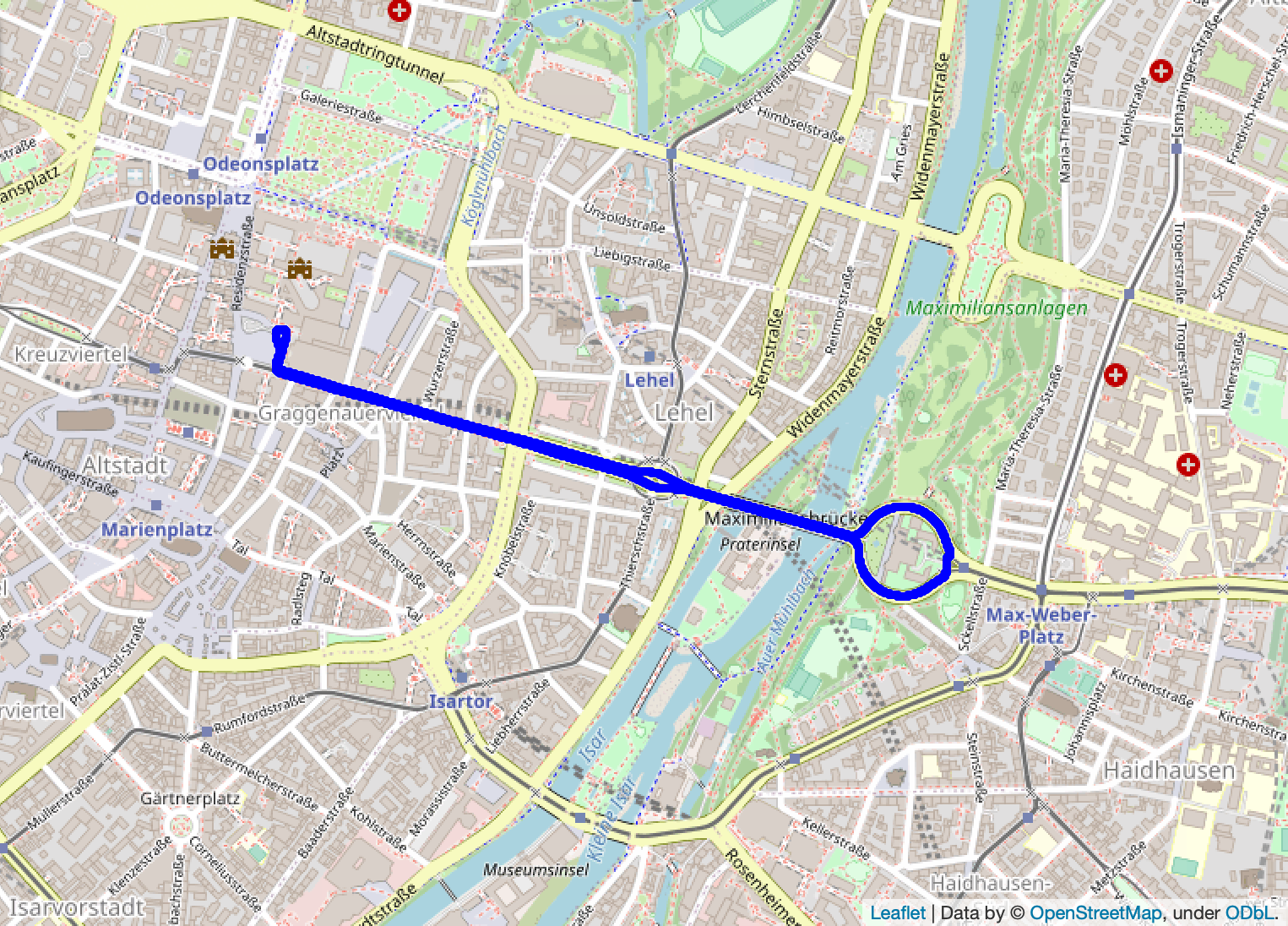}}
\qquad
\subfloat[][Parking Garage.]{\includegraphics[width=0.3\linewidth]{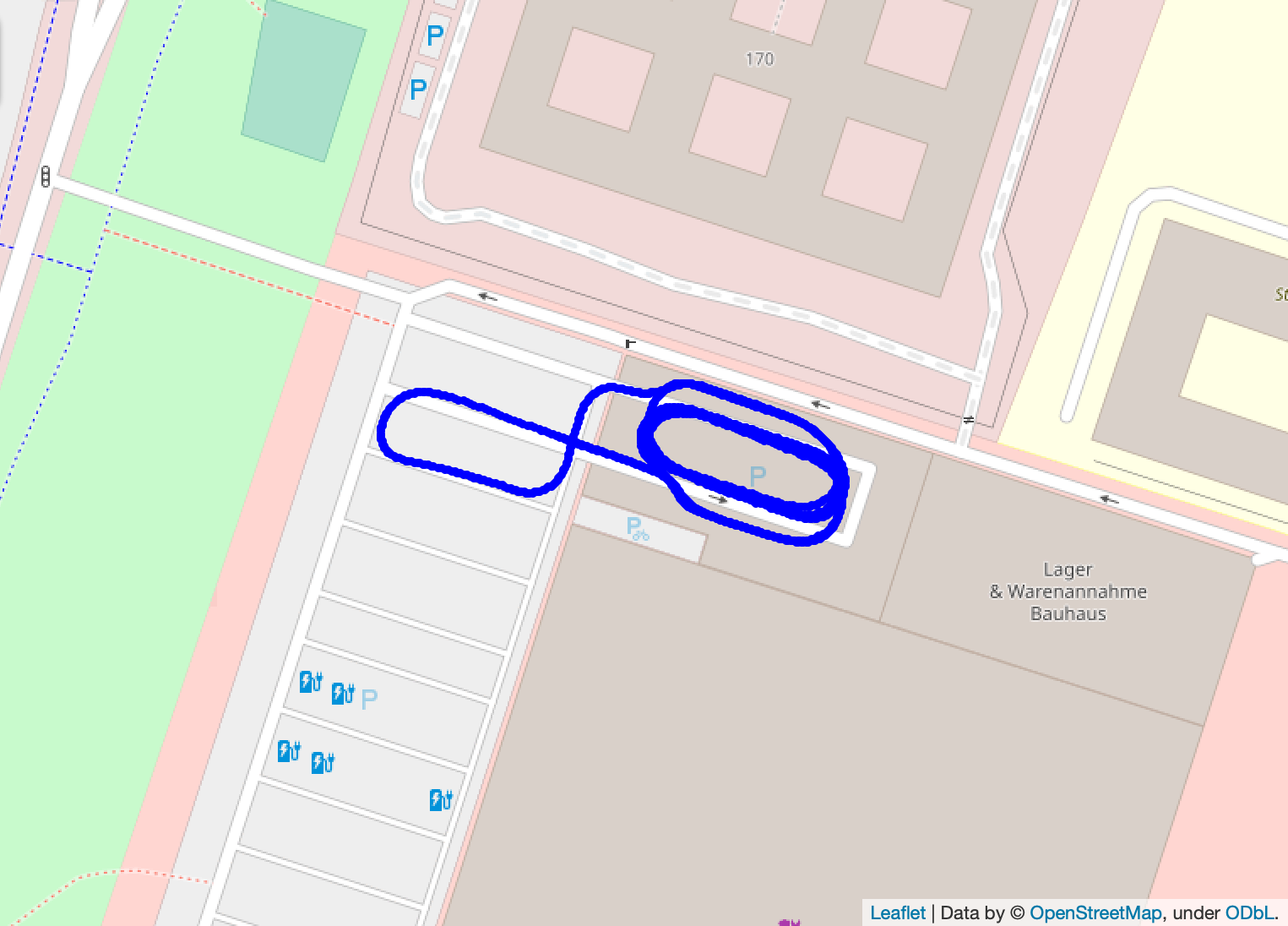}}

\caption{\textbf{Scenarios overview.} This figure shows all the covered scenarios of our benchmark dataset. We provide vastly different environments in and around the city of Munich, Germany.}%
\label{fig:app_all_scenarios}%
\end{figure*}

The \ac{vio} traces for each scenario are shown in Figure~\ref{fig:app_all_scenarios}. We provide reference poses and 3D models as sparse point clouds generated by our ground truth generation pipeline (\cf~Figure~\ref{fig:3d_models}) along with the corresponding raw image frames and raw \ac{imu} measurements. Figure~\ref{fig:3d_models_supplementary} shows an example of the optimized trajectory, which depicts the accuracy of the provided reference poses. Table~\ref{tab:4seasons_dataset_without_GT} shows all the sequences with withheld ground truth used for benchmarking.

\begin{figure*}
\centering
\includegraphics[width=0.8\linewidth]{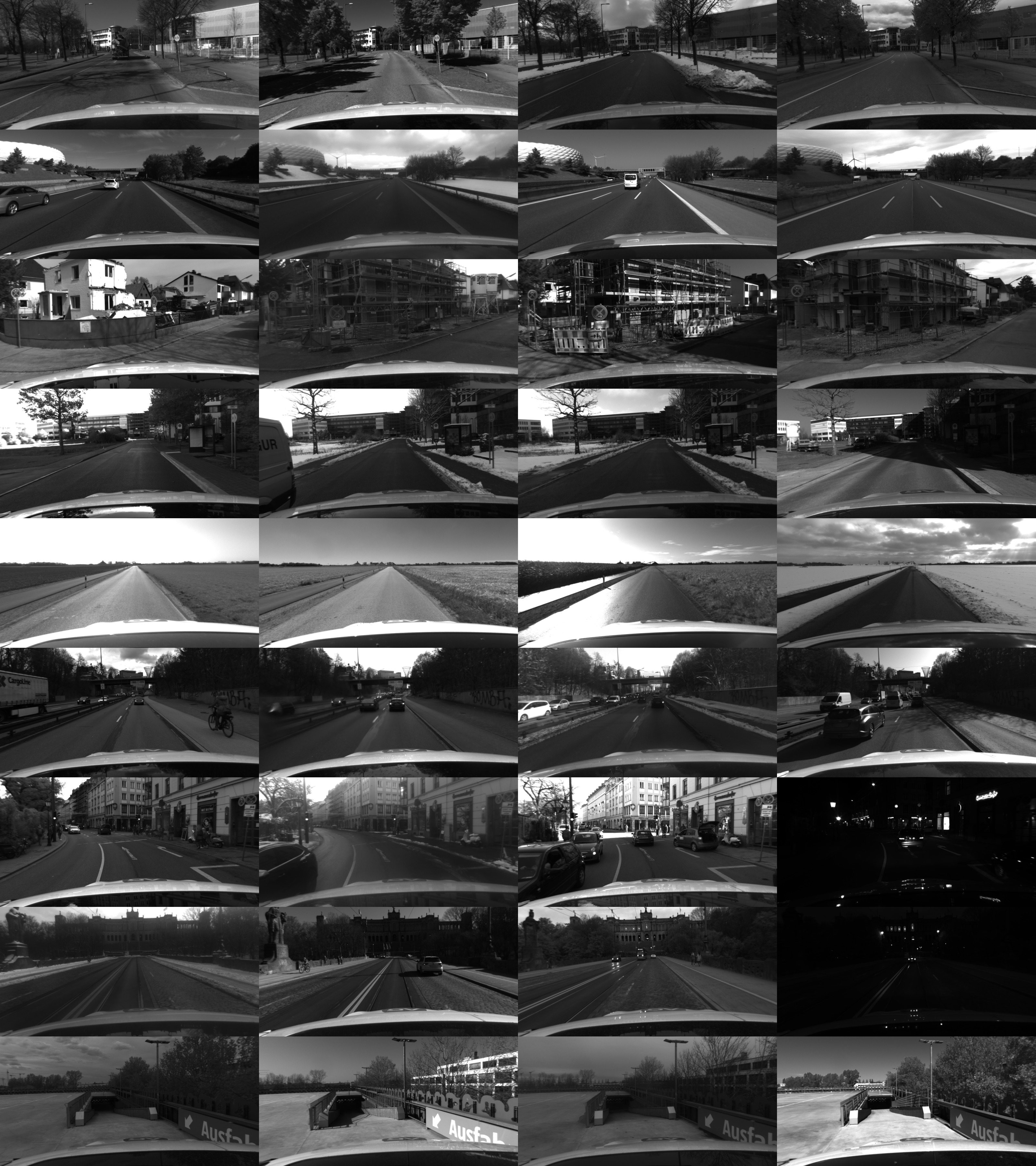}
\caption{\textbf{Dataset overview.} Example images from our benchmark dataset. First row: office loop, second row: highway, third row: neighborhood, fourth row: business campus, fifth row: countryside, sixth row: city loop, seventh row: old town, eighth row: maximilianeum, ninth row: parking garage. The figure illustrates the large appearance changes, occlusions, seasonal, and structural changes present in the data.}
\label{fig:dataset_overview_samples}
\end{figure*}

The benchmark dataset presents a challenge to current approaches to visual SLAM and long-term localization because it contains data from different seasons and weather conditions, as well as from different times of day, as shown in Figure~\ref{fig:dataset_overview_samples}. 

\subsection{Reference Pose Validation}\label{sec:ground_truth_validation}

\begin{figure}[t]
    \centering
    \includegraphics[width=0.8\linewidth]{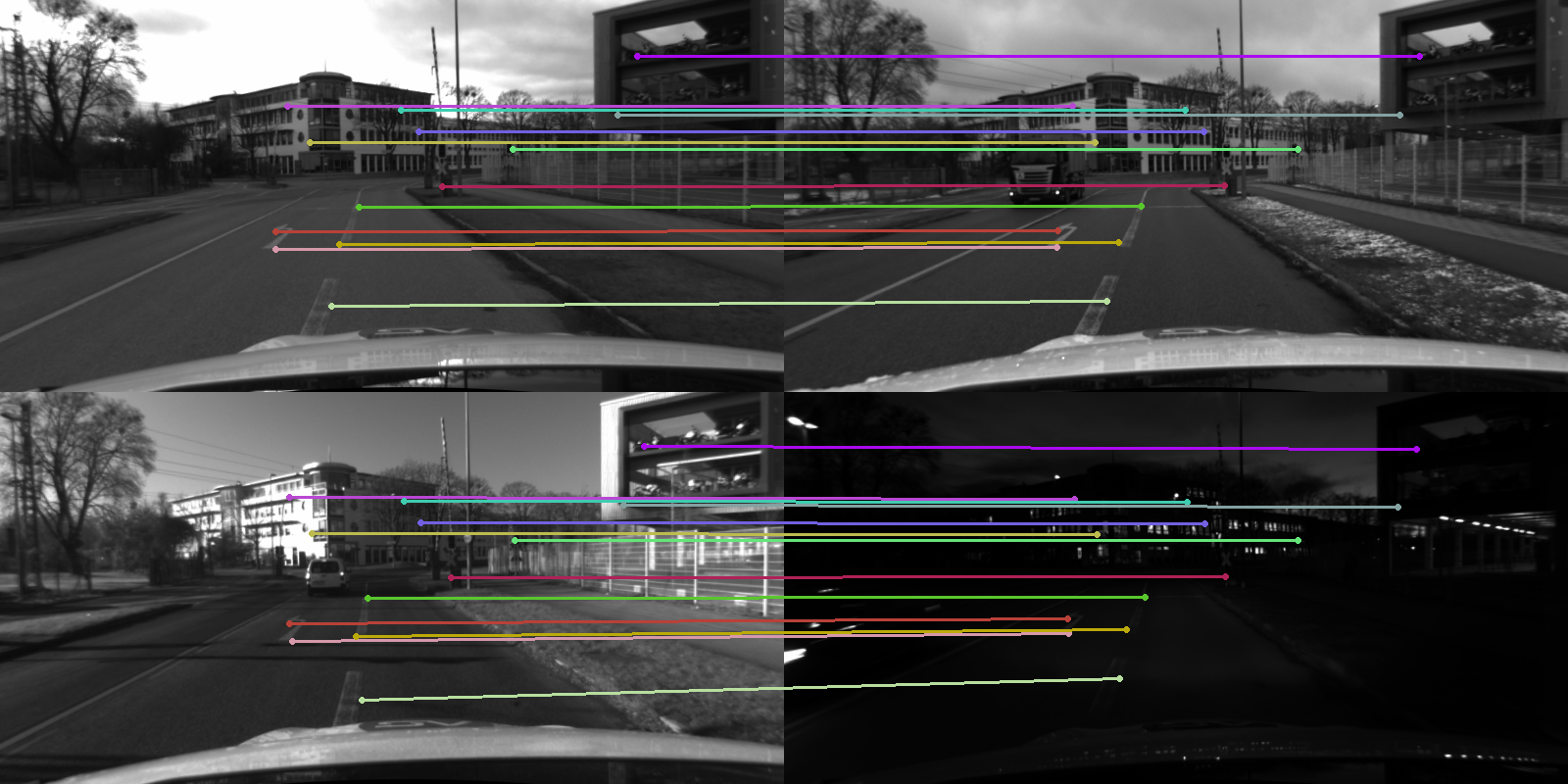}
    \caption{\textbf{Accurate pixel-wise correspondences, making cross-season training possible.} Qualitative assessment of the accuracy of our data collection and geometric reconstruction method for a sample of four different conditions (from top left in clockwise order: \emph{cloudy, snowy, night, sunny}) across the same scene. Each same colored point in the four images corresponds to the same geometric point in the world. The cameras corresponding to these images have different poses in the global frame of reference. Please note that the points are not matched, but rather a result of our accurate reference poses and geometric reconstruction.}
    \label{fig:corres_pts_img}
\end{figure}

The top part of Figure~\ref{fig:teaser} shows two overlaid point clouds from different runs across the same scene. Note that despite the weather and seasonal differences, the point clouds align very well. This shows that our reference poses are sufficiently accurate for benchmarking long-term localization. Furthermore, a qualitative assessment of the point-to-point correspondences is shown in Figure~\ref{fig:corres_pts_img}. The figure shows a subset of very accurate pixel-wise correspondences across different seasons (\emph{fall}/\emph{winter}) in the top and different illumination conditions (\emph{sunny}/\emph{night}) in the bottom. These point-to-point correspondences are a result of our up to centimeter-accurate global reference poses. This makes them suitable as training pairs for learning-based algorithms. Recently, there has been an increasing demand for pixel-wise cross-season correspondences, which are needed to learn dense feature descriptors~\citep{spencer2020same,Dusmanu2019CVPR,r2d2}. However, there is still a lack of datasets to satisfy this demand. The KITTI~\citep{Geiger2013IJRR} dataset does not provide cross-season data. The Oxford RobotCar Dataset~\citep{maddern20171} provides cross-season data, however, since the ground truth is not accurate enough, the paper does not recommend benchmarking localization and mapping approaches.

Recently, RobotCar Seasons~\citep{SattlerCVPR2018} was proposed to overcome the inaccuracy of the provided ground truth. However, similar to the authors of~\citep{spencer2020same}, we found that it is still challenging to obtain accurate cross-season pixel-wise matches due to pose inconsistencies. Furthermore, this dataset only provides images captured from three synchronized cameras mounted on a car, pointing to the rear-left, rear, and rear-right, respectively. Moreover, another limitation of the dataset is that it only provides relatively small segments and no long trajectories. Furthermore, a significant portion of it suffers from strong motion blur and low image quality.

\subsubsection{Pose Accuracy}\label{sec:pose_accuracy}
One potential limitation of our benchmark dataset is that we can only guarantee a certain pose accuracy when \ac{gnss} is available. Naturally, \ac{gnss} is unreliable in urban canyons or tunnels. Therefore, for the benchmark evaluation, we only consider poses as reference poses if \ac{gnss} is available and the observed standard deviation of the position is less than \SI{5}{\cm}. Please note that we only require accurate reference poses for the evaluation of visual localization. The evaluation of \ac{vo} is based on the accumulated drift over time, \ie it is only required that the start and end positions for each segment of a sequence are accurate. Furthermore, we provide quantitative measures of the quality of the maps. We report the percentage of accurate reference poses for each trajectory. Moreover, we report the overall map accuracy in terms of horizontal RMSE between the \ac{gnss} poses and the refined poses after pose graph optimization.

The percentage of accurate poses for each test sequence can be seen in Table~\ref{tab:4seasons_dataset_without_GT} and Table~\ref{tab:4seasons_dataset_with_GT} for the training sequences. For a qualitative visual analysis, we show accurate pixel-wise correspondences in Figure~\ref{fig:corres_pts_img}, indicating that the reference poses are sufficiently accurate. We do not claim that our poses are consistently centimeter-accurate, however, by analyzing the map accuracy we can assure the quality of the poses used for benchmarking.

\subsection{Data Source}
We release (distorted \& undistorted) 8-bit grayscale images, \ac{imu} measurements, and sensor calibration, including the calibration sequences, for all sequences (training and testing). In addition, \ac{rtk} \ac{gnss} measurements, in NMEA format, \ac{vo} point clouds, and reference poses are released only for training sequences. For the testing sequences, such data is withheld for evaluation. Moreover, we specify the distance between the refined reference poses and the raw \ac{rtk} \ac{gnss} measurements. 

\section{Benchmark Tasks}\label{sec:benchmark_tasks}
In this section, we define the benchmark evaluation metrics, tasks, and their evaluation protocols for \acl{vo}, global place recognition, and map-based visual localization. Visual localization consists of retrieving the \ac{6dof} pose of a query within an existing 3D model and can be interpreted as a two-step approach. First, global image retrieval is performed to obtain a rough estimate of the query pose \wrt a map. Second, local feature matching is used to refine the pose estimate. 

For the evaluation, in each task, we consider a set of estimated \ac{6dof} poses $\mathbf{T}_{i}^\text{est} \in \mathrm{SE}(3)$, as well as a set of reference, poses $\mathbf{T}_{i}^\text{ref} \in \mathrm{SE}(3)$. While the reference poses are always defined \wrt a global world frame, the estimated poses are defined either \wrt the same global world frame (for global place recognition and map-based visual localization) or to a selected local frame\footnote{Can be for instance the camera frame of the first recorded left camera image.} (\acl{vo}).

\subsection{Visual Odometry in Challenging Conditions}\label{subsec:odometry_benchmark}
\Acl{vo} aims to accurately estimate the relative \ac{6dof} camera pose based on recorded images. To benchmark the task of \ac{vo} there already exists various datasets~\citep{Geiger2012CVPR,sturm12iros,engel2016monodataset}. All of these existing datasets consist of sequences recorded at rather homogeneous conditions (indoors, or sunny/overcast outdoor conditions). However, methods specially developed for autonomous driving use cases must perform robustly under almost any condition. We believe that the proposed benchmark will contribute to improving the performance of \ac{vo} under diverse weather and lighting conditions in an automotive environment. Therefore, instead of replacing existing benchmarks and datasets, we aim to provide an extension that is more focused on challenging conditions in autonomous driving. As we provide frame-wise accurate poses for large portions of the sequences, metrics well known from other benchmarks like \ac{ate} or \ac{rpe}~\citep{Geiger2012CVPR,sturm12iros} are also applicable to our data.

\subsubsection{Evaluation Metrics}
Similar to previous benchmarks, the main accuracy measure we are interested in is the \ac{rpe}. In general, the \ac{rpe} is split up into a translational and a rotational error. However, another component we are interested in is the scale error. One may argue that, especially for stereo approaches, scale errors are marginal and therefore not relevant. Nevertheless, our experience is different. We observe that quite significant scale errors and drift can occur when performing stereo \ac{vo} and \ac{slam} in automotive environments. This can be caused either by the miss-calibration of the cameras, by the structure of the scene but also by algorithm-specific design choices like the type of keypoint detector, \etc. Since the sensor setup has a limited stereo baseline, parallaxes (\ie pixel disparities) for far object points are vanishing. This means that, even for stereo approaches, the scale becomes non-observable if no close static objects are present in the scene. Increasing the stereo baseline, however, could reduce the rigidity of the sensor setup. We believe that it is very valuable to conduct further research on stereo \ac{vo} and \ac{slam} methods which explicitly consider the depth uncertainties created by the length of the stereo baseline.

Since in automotive use cases, the scale can always be observed based on a reference system, like wheel ticks, \ac{gnss} or a reference map, we consider only relative errors (drifts) in scale, translation, and rotation in the proposed benchmark. Therefore, before evaluation, a global scale alignment is performed for the entire trajectory with respect to the reference trajectory.

For the proposed \ac{vo} benchmark all evaluation metrics are defined based on the estimated relative pose $\mathbf{T}_{ij}^\text{est} \in \mathrm{SE}(3)$ between two frames $i$ and $j$ and its corresponding reference pose $\mathbf{T}_{ij}^\text{ref} \in \mathrm{SE}(3)$ with:
\begin{align}
\mathbf{T}_{ij}^\text{ref} &=  \left(\mathbf{T}_{i}^\text{ref}\right)^{-1}\mathbf{T}_{j}^\text{ref} \quad \text{and} \quad \nonumber\\
\mathbf{T}_{ij}^\text{est} &= \left(\mathbf{T}_{i}^\text{est}\right)^{-1}\mathbf{T}_{j}^\text{est}.
\end{align}
For a pair of frames ($i$, $j$) for which reference poses are available, we calculate the relative translational error $\epsilon_{ij}^t$, rotational error $\epsilon_{ij}^r$, and scale error $\Tilde{\epsilon}_{ij}^s$ as given in Equations~\eqref{eq:trans_error_odom} to~\eqref{eq:scale_error_odom}.
\begin{align}
\epsilon_{ij}^t &=  \frac{\|\mathbf{t}_{ij}^{\text{ref}} - \mathbf{t}_{ij}^{\text{est}}\|_2}{d_{ij}} \label{eq:trans_error_odom}\\
\epsilon_{ij}^r &= \frac{\arccos\left(\frac{1}{2} \left(\operatorname{trace}{((\mathbf{R}_{ij}^\text{ref})^{-1}\mathbf{R}_{ij}^\text{est})} - 1\right)\right)}{d_{ij}} \label{eq:rot_error_odom}\\
\Tilde{\epsilon}_{ij}^{s} &= \frac{\left\|\mathbf{t}_{ij}^{\text{est}}\right\|_2}{\left\|\mathbf{t}_{ij}^{\text{ref}}\right\|_2} \label{eq:scale_error_odom}
\end{align}
From $\Tilde{\epsilon}_{ij}^{s}$ one obtains the final relative scale error as $\epsilon_{ij}^{s} = \max[\Tilde{\epsilon}_{ij}^{s}, (\Tilde{\epsilon}_{ij}^{s})^{-1}]$. The parameter $d_{ij}$ defines the reference path length between the two poses $\mathbf T_i^\text{ref}$ and $\mathbf T_j^\text{ref}$.

Meaningful metrics are obtained by extracting all possible sub-segments of length \SI{100}{\m}, \SI{200}{\m}, \SI{400}{\m}, \SI{600}{\m}, \SI{800}{\m}, and \SI{1000}{\m} from a trajectory and calculating the relative poses between the first and last frame of each sub-segment.
Furthermore, for trajectory segments where no \ac{gnss} measurements are available for more than \SI{1000}{\m} (\eg in tunnels, garages, or urban canyons), also the relative poses of such an entire stretch are taken into account. This allows us to also consider challenging scenarios like tunnels and the transition from bright to dark in the benchmark. Using sub-segments of different lengths for evaluation is inspired by the KITTI benchmark~\citep{Geiger2012CVPR} and allows capturing both short and long-term accuracy of \ac{vo} algorithms.

To obtain single number metrics for every sequence, we consider the visual \ac{vo} successful if the errors are within certain positional, rotational, and scale bounds. We define three intervals by varying the thresholds: \textbf{high precision} (\SI{0.5}{\percent}, \SI{0.005}{deg/\m}, 1.005 (multiplier)), \textbf{medium precision} (\SI{1}{\percent}, \SI{0.01}{deg/\m}, 1.01 (multiplier)), and \textbf{coarse precision} (\SI{2}{\percent}, \SI{0.02}{deg/\m}, 1.02 (multiplier)).

While the translational error is the most meaningful metric to evaluate \ac{vo} algorithms, the rotational error, and scale error still give valuable insight into the specific behavior of a certain approach.

\subsection{Global Place Recognition}\label{subsec:global_image_retrieval}
Global place recognition refers to the task of retrieving the most similar database image given a query image~\citep{lowry2015visual}. To improve the searching efficiency and the robustness against different weather conditions, tremendous progress on global descriptors~\citep{jegou2010aggregating,arandjelovic2013all,angeli2008fast,galvez2012bags} has been seen. For the localization pipeline, visual place recognition serves as the initialization step to the downstream local pose refinement by providing the most similar database images as well as the corresponding global poses. Due to the advent of deep neural networks~\citep{simonyan2014very,krizhevsky2012imagenet,he2016deep,szegedy2015going}, methods aggregating deep image features are proposed and have shown advantages over classical methods~\citep{arandjelovic2016netvlad,gordo2016deep,radenovic2018fine,tolias2015particular}.

\begin{figure}[t]
    \centering
    \includegraphics[width=0.8\linewidth]{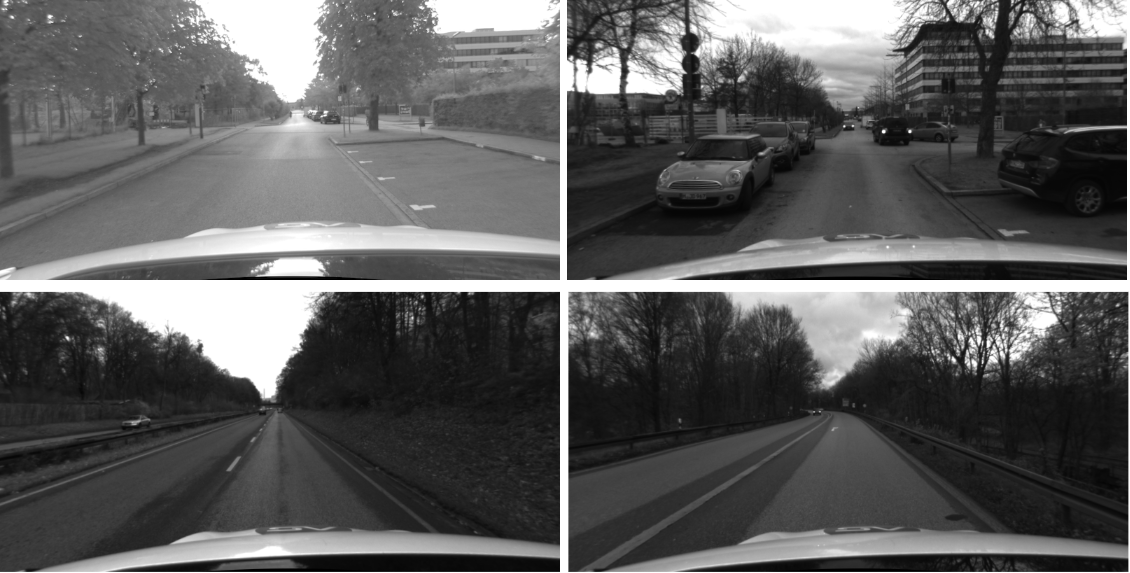}
    \caption{\textbf{Challenging scenes for global place recognition.} Top: two pictures share the same location with different appearances. Bottom: two pictures have a similar appearance but are taken at different locations.}
    \label{fig:vpr}
\end{figure}

The proposed dataset is challenging for global place recognition since it contains not only cross-season images that have different appearances with a similar geographical location but also intra-season images which share similar appearances but with different locations. This results in mainly two different types: images taken at the same place, but look different, or images taken at different places but look similar. Figure~\ref{fig:vpr} depicts example pairs of these scenarios. 

\subsubsection{Evaluation Metrics}
We follow the standard metric widely used for global place recognition~\citep{arandjelovic2016netvlad,arandjelovic2013all,sattler2012image,gordo2016deep}, namely the recall at top $N$ retrievals with a certain range bound as the positive threshold. Specifically, one query image is considered to be correctly localized if at least one of the top $N$ retrieved images is within a certain translational (in meters) and a certain rotational (in degrees) bound with respect to the ground-truth location of the query image. The translational error $\epsilon^t$ is measured as the Euclidean distance:
\begin{align}
    \epsilon^t = \|\mathbf{t}^{\text{ref}} - \mathbf{t}^{\text{est}}\|_2
\label{eq:trans_error_reloc}
\end{align}
between the reference $\mathbf{t}^{\text{ref}}$ and estimated $\mathbf{t}^{\text{est}}$ camera positions. The rotational error $\epsilon^r$ is measured as an angle in degrees (following~\citep{hartley2013rotation}) by calculating:
\begin{align}
    \epsilon^r = \arccos\left(\frac{1}{2} \left(\operatorname{trace}{((\mathbf{R}^\text{ref})^{-1}\mathbf{R}^\text{est})} - 1\right)\right),
\label{eq:rot_error_reloc}
\end{align}
where $\mathbf{R}^{\text{ref}}$, and $\mathbf{R}^{\text{est}}$ denote the reference and estimated camera rotation matrices. In the evaluation of global place recognition, we calculate the recalls under different threshold settings: by fixing $N$ and changing the range bound, or by fixing the range bound and changing $N$. We will describe the specific settings in Section~\ref{sec:exp_gpr}.

\subsection{Map-Based Visual Localization}\label{subsec:reloc_tracking}
Map-based visual localization refers to the task of locally refining the \ac{6dof} pose between reference images and images from a query sequence. In contrast to wide-baseline stereo matching, for map-based visual localization, it is also possible to utilize the sequential information of the sequence. This allows estimating depth values by running a standard \ac{vo} method. Those depth estimates can then be used to improve the tracking of the individual localization candidates.

In contrast to global place recognition which only uses 2D images and no other information, this task allows the use of a globally-consistent 3D reconstruction of the reference scene. In this task, we assume to know the mapping between reference and query samples and only focus on the local pose refinement task. In practice, this mapping can be found using image retrieval techniques as described in Section~\ref{subsec:global_image_retrieval} or by using \ac{gnss} measurements as a coarse initialization if available.

Accurately localizing in a pre-built map is a challenging problem, especially if the visual appearance of the query sequence significantly differs from the base map. This makes it extremely difficult, especially for vision-based systems, since the localization accuracy is often limited by the discriminative power of feature descriptors. Our proposed dataset allows evaluating visual localization across multiple types of weather conditions and diverse scenes, ranging from urban to countryside driving. Furthermore, our up to centimeter-accurate reference poses allow us to create more strict evaluation settings with an increased level of difficulty. This allows us to determine the limitations and robustness of current state-of-the-art methods.

\subsubsection{Evaluation Metrics}
For evaluation, we measure the translational and rotational error of any method between the estimated and the reference pose. Please refer to Equation~\eqref{eq:trans_error_reloc} and Equation~\eqref{eq:rot_error_reloc} for the definitions of the translational and rotational error, respectively.

We consider the localization successful if a query image is localized within certain positional (in meters) and rotational (in degrees) bounds with respect to their reference pose. We define three localization intervals by varying the thresholds: \textbf{high precision} (\SI{0.1}{\m}, \SI{1}{\degree}), \textbf{medium precision} (\SI{0.25}{\m}, \SI{2}{\degree}), and \textbf{coarse precision} (\SI{1}{\m}, \SI{5}{\degree}). 

\section{Experimental Evaluation}\label{sec:experimental_results}
In this section, we evaluate the current state-of-the-art baseline methods for each of the three provided benchmarks (\acl{vo}, global place recognition, and map-based visual localization) to demonstrate the diversity and challenges of the benchmark. We will establish an open leaderboard for the benchmark to compare different methods upon publication. This allows the reproduction of the baseline results for every user. Furthermore, we will set up a server for an automatic evaluation of the results on the withheld test set.

\subsection{Visual Odometry in Challenging Conditions}\label{sec:vo_results}
We provide results for state-of-the-art baseline stereo and stereo-inertial odometry and \ac{slam} approaches. The methods provided as baselines are classical geometric approaches. Nevertheless, we strongly encourage researchers to evaluate learning-based methods on our benchmark as well. In particular, we provide results for the following stereo and stereo-inertial \ac{vo} methods: ORB-SLAM3\footnote{\url{https://github.com/UZ-SLAMLab/ORB_SLAM3}}~\citep{ORBSLAM3_2020} and Basalt\footnote{\url{https://gitlab.com/VladyslavUsenko/basalt}}~\citep{usenko19nfr}.

\begin{table*}[t]
\begin{center}
\ra{1.3}
\caption{\textbf{\Acl{vo} results on \emph{known scenarios} from the 4Seasons benchmark.} This table shows the evaluation results of state-of-the-art baseline methods on the \ac{vo} benchmark. The best-performing results are in bold. The results are shown in terms of the percentage of high / medium / coarse precision.}
\label{tab:vo_benchmark_results_known}
\resizebox{\textwidth}{!}{
\begin{tabular}{@{}lccccccccc|c@{}}
\toprule
\bfseries Method & \bfseries office\_loop\_2\_test & \bfseries office\_loop\_3\_test & \bfseries neighborhood\_2\_test & \bfseries business\_campus\_1\_test & \bfseries countryside\_2\_test & \bfseries city\_loop\_2\_test & \bfseries old\_town\_2\_test & \bfseries old\_town\_3\_test & \bfseries parking\_garage\_1\_test & \bfseries Average \\
\midrule
Basalt~\citep{usenko19nfr} (stereo) & 9.1 / 65.7 / 96.7 & 6.3 / 53.0 / 94.4 & 4.2 / 21.5 / 70.1 & 2.3 / 28.5 / 71.5 & 7.7 / 38.3 / 77.6 & 11.4 / 43.8 / 72.3 & 5.6 / 31.1 / 78.0 & 1.3 / 9.0 / 37.4 & 0.0 / 0.0 / 33.3 & 5.3 / \textbf{32.3} / \textbf{70.2}\\
\midrule
Basalt~\citep{usenko19nfr} (stereo-inertial) & 3.3 / 35.0 / 92.0 & 2.1 / 20.9 / 80.8 & 3.5 / 23.6 / 72.9 & 11.3 / 59.0 / 95.7 & 16.4 / 48.5 / 88.8 & 23.1 / 59.2 / 88.2 & 0.0 / 0.0 / 0.0 & 1.5 / 15.4 / 42.6 & 0.0 / 11.1 / 55.6 & 6.8 / 30.3 / 68.5\\
\midrule
ORB-SLAM3~\citep{ORBSLAM3_2020} (stereo) & 16.8 / 65.3 / 94.9 & 1.4 / 24.0 / 82.2 & 4.9 / 55.6 / 95.8 & 3.9 / 42.2 / 82.8 & 5.8 / 41.4 / 76.6 & 1.2 / 12.6 / 49.3 & 0.8 / 17.7 / 57.0 & 0.3 / 1.0 / 2.8 & 0.0 / 22.2 / 77.8 & 3.9 / 31.3 / 68.8\\
\midrule
ORB-SLAM3~\citep{ORBSLAM3_2020} (stereo-inertial) & 7.3 / 33.6 / 84.7 & 2.1 / 15.7 / 50.2 & 13.2 / 44.4 / 84.0 & 19.9 / 64.8 / 91.0 & 2.9 / 11.4 / 42.6 & 26.1 / 59.2 / 77.9 & 12.9 / 43.3 / 87.1 & 0.0 / 0.0 / 0.0 & 0.0 / 11.1 / 44.4 & \textbf{9.4} / 31.5 / 62.4\\
\bottomrule
\end{tabular}
}
\end{center}
\end{table*}
\begin{table}[t]
\begin{center}
\ra{1.3}
\caption{\textbf{\Acl{vo} results on \emph{unknown scenarios} from the 4Seasons benchmark.} This table shows the evaluation results of state-of-the-art baseline methods on the \ac{vo} benchmark. The best-performing results are in bold. The results are shown in terms of the percentage of high / medium / coarse precision.}
\label{tab:vo_benchmark_results_unknown}
\resizebox{\columnwidth}{!}{
\begin{tabular}{@{}lcccc|c@{}}
\toprule
\bfseries Method & \bfseries highway\_1\_test & \bfseries highway\_2\_test & \bfseries maximilianeum\_1\_test & \bfseries maximilianeum\_2\_test & \bfseries Average\\
\midrule
Basalt~\citep{usenko19nfr} (stereo) & 9.4 / 32.0 / 63.1 & 10.3 / 29.5 / 52.7 & 35.1 / 75.3 / 91.8 & 1.2 / 9.8 / 38.2 & 14.0 / 36.6 / 61.4\\
\midrule
Basalt~\citep{usenko19nfr} (stereo-inertial) & 32.3 / 68.6 / 85.8 & 21.3 / 49.2 / 71.2 & 34.0 / 69.6 / 94.3 & 0.0 / 6.9 / 27.2 & \textbf{21.9} / \textbf{48.6} / \textbf{69.6}\\
\midrule
ORB-SLAM3~\citep{ORBSLAM3_2020} (stereo) & 0.6 / 3.9 / 22.1 & 0.0 / 0.0 / 0.0 & 0.0 / 19.6 / 56.7 & 0.0 / 0.0 / 0.0 & 0.2 / 5.9 / 19.7\\
\midrule
ORB-SLAM3~\citep{ORBSLAM3_2020} (stereo-inertial) & 10.3 / 29.6 / 48.6 & 12.9 / 25.1 / 68.3 & 26.3 / 60.8 / 79.4 & 10.4 / 39.3 / 56.1 & 15.0 / 38.7 / 63.1\\
\bottomrule
\end{tabular}
}
\end{center}
\end{table}

\begin{figure*}[t]
\centering
    \includegraphics[width=0.8\linewidth]{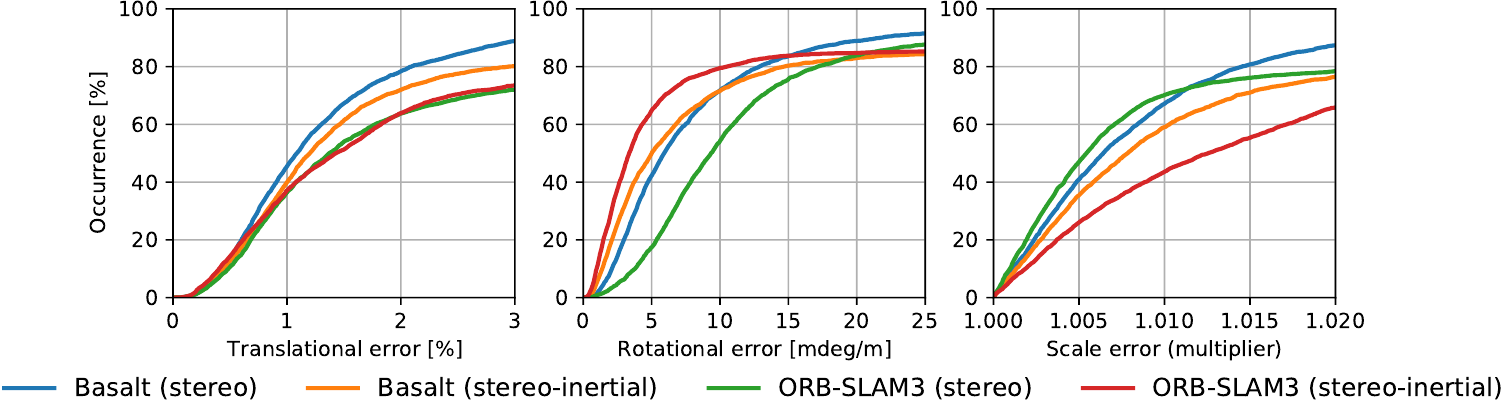}
    \caption{\textbf{Performance of state-of-the-art baseline \acl{vo} methods on \emph{known scenarios} from the 4Seasons benchmark.} The figure shows the translational error (in $\%$), rotational error (in $\text{mdeg}/\text{m}$), and scale error (multiplier).}
    \label{fig:odom_eval_known_scenes}
\end{figure*}

\begin{figure*}[t]
\centering
    \includegraphics[width=0.8\linewidth]{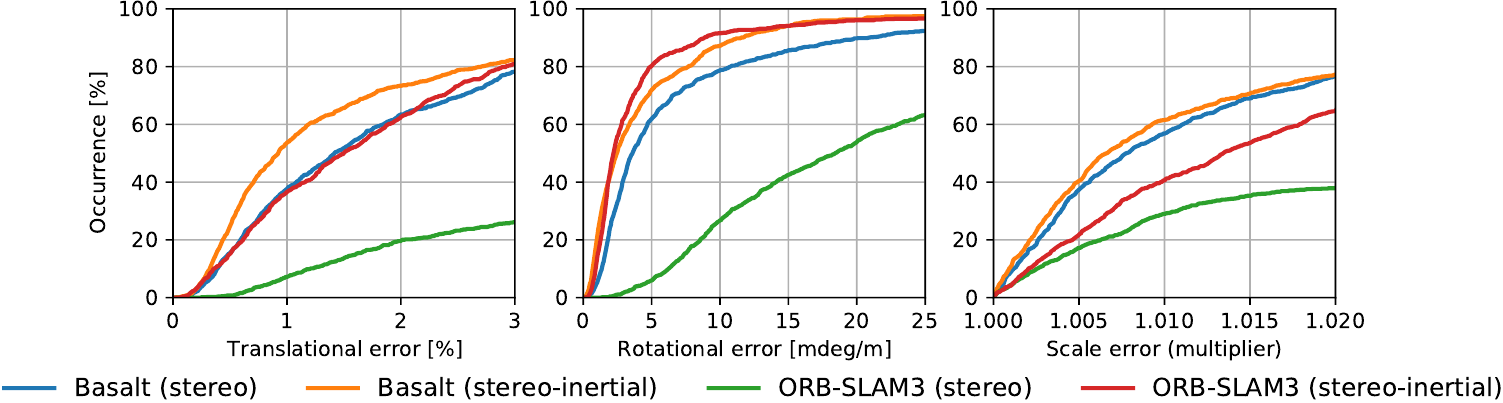}
    \caption{\textbf{Performance of state-of-the-art \acl{vo} methods on \emph{unknown scenarios} from the 4Seasons benchmark.} The figure shows the translational error (in $\%$), rotational error (in $\text{mdeg}/\text{m}$), and scale error (multiplier).}
    \label{fig:odom_eval_unknown_scenes}
\end{figure*}

The provided \ac{vo} benchmark is divided into two sets of evaluation sequences: \emph{unknown scenarios} and \emph{known scenarios}. \emph{Unknown scenarios} consist only of scenarios, for which no sequences at all are provided in the training set. Namely, these are the scenarios \emph{Highway} and \emph{Maximilianeum}. \emph{Known scenarios} are those scenarios for which are also sequences provided in the training set. While this is irrelevant for pure geometric approaches, we believe that this separation will be important to evaluate the generalization capabilities of learning-based approaches. Table~\ref{tab:vo_benchmark_results_known} shows the evaluation results on the individual sequences of the benchmark for \emph{known scenarios}. Figure~\ref{fig:odom_eval_known_scenes} shows the results across all sequences corresponding to the \emph{known scenarios} in cumulative error plots. Table~\ref{tab:vo_benchmark_results_unknown} shows the evaluation results on the individual sequences of the benchmark for \emph{unknown scenarios}. Figure~\ref{fig:odom_eval_unknown_scenes} shows the results across all sequences corresponding to the \emph{unknown scenarios} in cumulative error plots.

From Table~\ref{tab:vo_benchmark_results_known} and~\ref{tab:vo_benchmark_results_unknown} one can observe that all evaluated methods perform significantly worse on the \emph{unknown scenarios}. Nevertheless, this is mainly due to the challenging conditions, which are on one side highway sequences with high speed and sudden lighting changes under bridges as well as inner city night sequences.

\begin{figure*}[t]
\centering
\subfloat[][Afternoon.]{\includegraphics[width=0.6\linewidth]{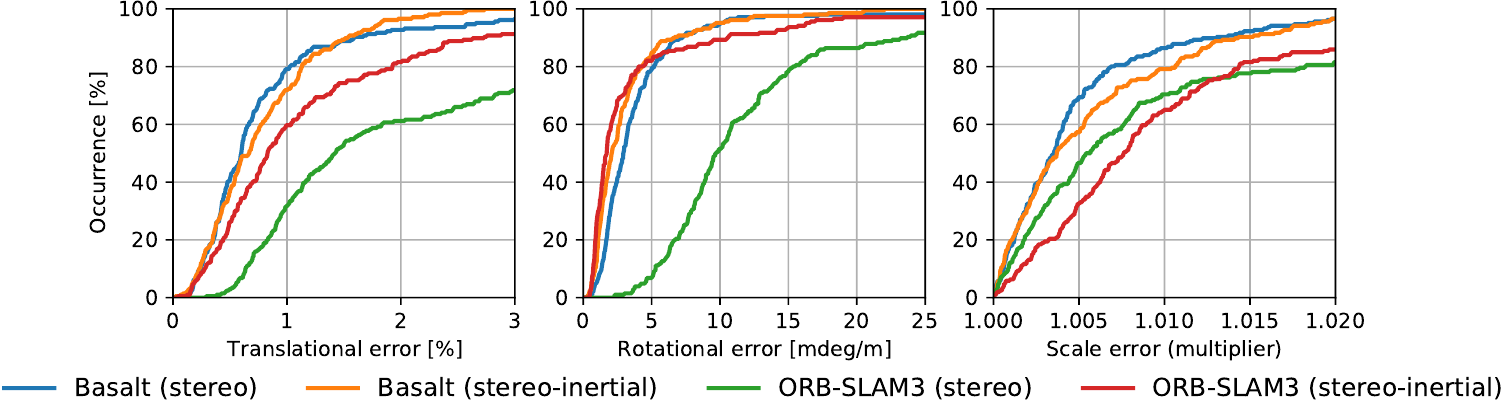} \quad \includegraphics[width=0.3\linewidth]{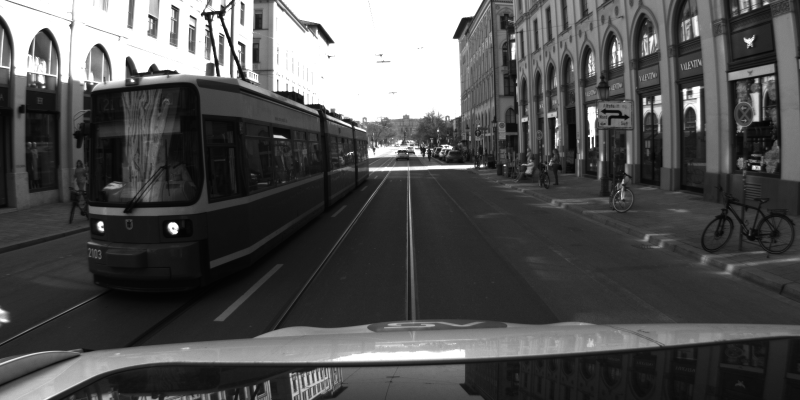}}\\
\subfloat[][Night.]{\includegraphics[width=0.6\linewidth]{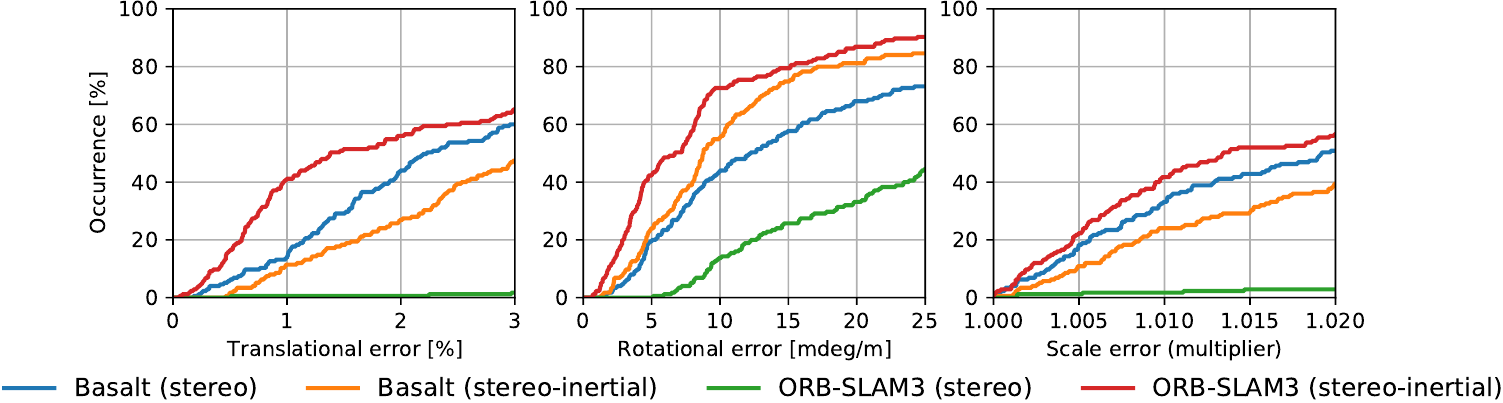} \quad \includegraphics[width=0.3\linewidth]{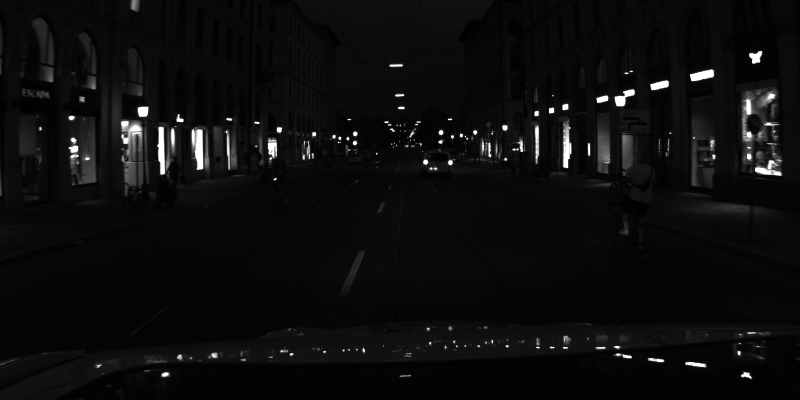}}
\caption{\textbf{Comparison of \acl{vo} performance for afternoon and night.} The figure shows the performance for different state-of-the-art baseline \ac{vo} algorithms on the same route for afternoon and night conditions. One can observe a significant drop in performance when going from day to night due to less visible landmarks.}
\label{fig:odom_eval_day_night}
\end{figure*}

\begin{figure*}[t]
\centering
\subfloat[][Sunny.]{\includegraphics[width=0.6\linewidth]{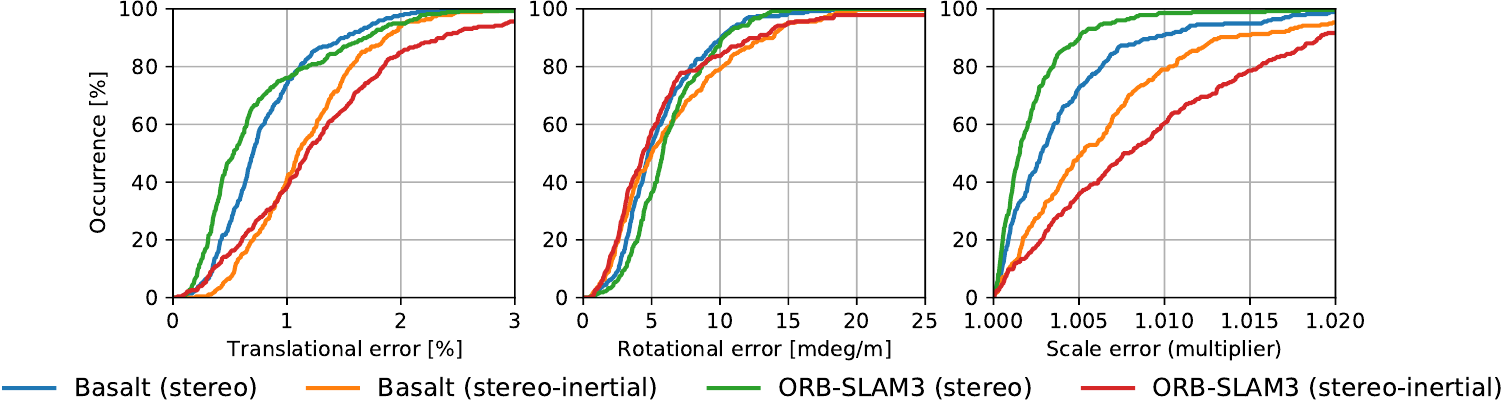} \quad \includegraphics[width=0.3\linewidth]{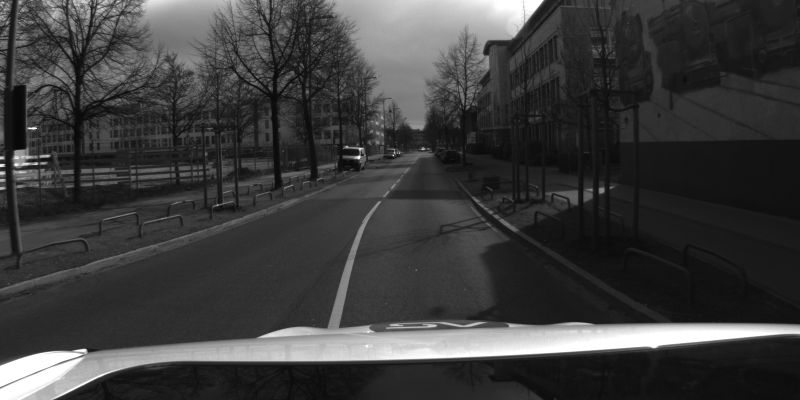}}\\
\subfloat[][Cloudy.]{\includegraphics[width=0.6\linewidth]{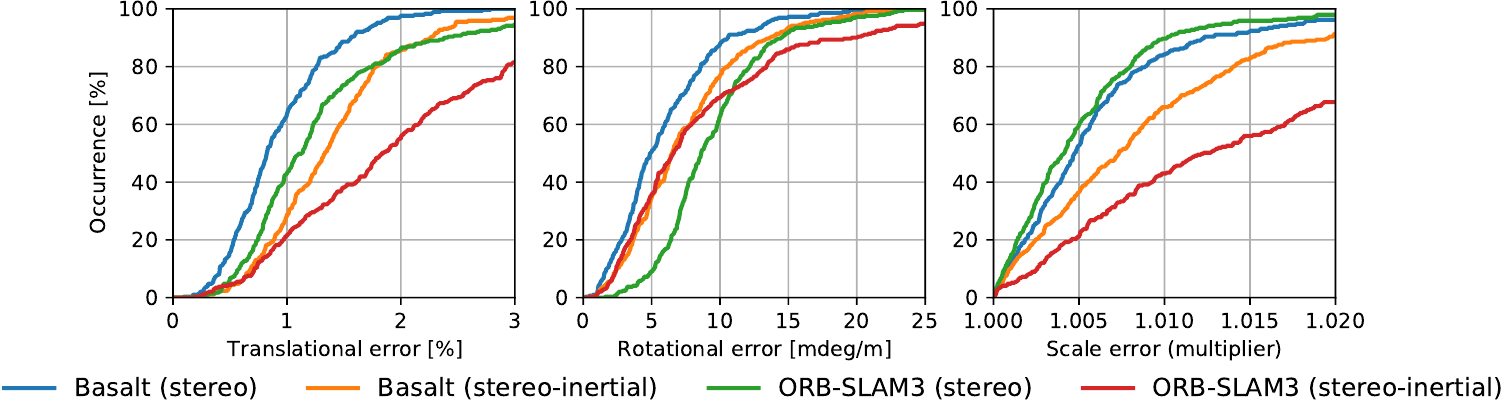} \quad \includegraphics[width=0.3\linewidth]{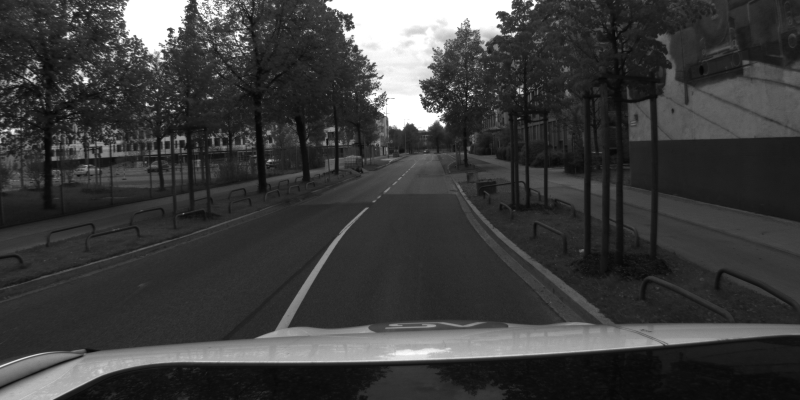}}
\caption{\textbf{Comparison of \acl{vo} performance for sunny and cloudy.} The figure shows the performance for different state-of-the-art baseline \ac{vo} algorithms on the same route for sunny and cloudy conditions. Across all algorithms, one can observe improved performance during sunny weather conditions. A likely reason for this is the presence of more static feature points caused by shadows, especially on the road. This can be seen on the right side images, where much more texture is on the road for the sunny than for the cloudy conditions.}
\label{fig:odom_eval_sunny_overcast}
\end{figure*}

While the results above provide average numbers across all sequences of the benchmark, we provide in Figure~\ref{fig:odom_eval_day_night} and~\ref{fig:odom_eval_sunny_overcast} side-by-side the results for identical scenarios but for different conditions, respectively.

Figure~\ref{fig:odom_eval_day_night} provides \ac{vo} results on the \emph{Maximilianeum} scenario in the afternoon (maximilianeum\_1\_test) and at night (maximilianeum\_2\_test), respectively. As one could expect, there is a significant drop in performance when going from day to night due to less visible landmarks. Nevertheless, it is interesting to observe that ORB-SLAM3 (with \ac{imu}) can perform better during the night than Basalt (with \ac{imu}). A reason might be that ORB-SLAM3 is using feature matching to find point correspondences, while Basalt is relying on optical flow. This correlation cannot be observed when running without \ac{imu}, where ORB-SLAM3 is failing. However, especially during the night and without \ac{imu}, the task becomes inordinately more difficult.

Figure~\ref{fig:odom_eval_sunny_overcast} provides performance comparisons between a sunny (office\_loop2\_test) and a cloudy (office\_loop3\_test) condition on the \emph{Office Loop} scenario. Across all algorithms, one can observe improved performance during sunny weather conditions. A likely reason for this is the presence of more static feature points caused by shadows, especially on the road. This can be seen on the right side images in Figure~\ref{fig:odom_eval_sunny_overcast}, where much more texture is on the road for the sunny than for the cloudy conditions.

While the evaluated methods show all-in-all good performance in good weather and lighting conditions, we believe that our dataset and benchmark will contribute to improving the performance in conditions with fewer and unreliable feature points. The results show that the proposed benchmark is highly challenging and still provides room for improving state-of-the-art \ac{vo} algorithms.


\subsection{Global Place Recognition}\label{sec:exp_gpr}
We evaluate the current state-of-the-art baseline deep image descriptors methods including NetVLAD\footnote{\url{https://github.com/cvg/Hierarchical-Localization}}~\citep{arandjelovic2016netvlad} pretrained on Pittsburgh30k~\citep{torii2013visual}, Deep Image Retrieval (DIR)\footnote{\url{https://github.com/naver/deep-image-retrieval}}~\citep{GARL17,RARS19} (aka AP-GeM) trained on the Landmarks dataset~\citep{babenko2014neural}, and CNN Image Retrieval (CIR)\footnote{\url{https://github.com/filipradenovic/cnnimageretrieval-pytorch}}~\citep{radenovic2016cnn,radenovic2018fine} trained on the dataset derived from~\citep{schonberger2015single}. For each scenario of the 4Seasons benchmark, we use a predefined recording as the reference map and a predefined recording as the query map. Note that we leave out the \emph{Business Campus} scenario for global place recognition.

\begin{figure}[t]
\centering
    \includegraphics[width=0.8\linewidth]{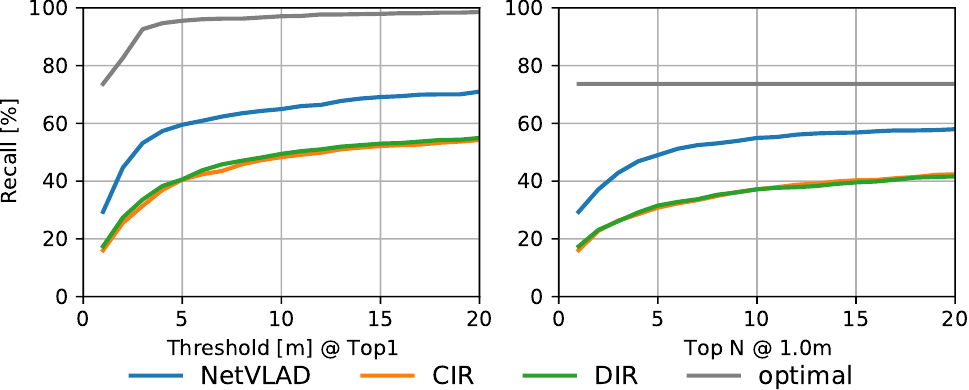}
    \caption{\textbf{Performance of state-of-the-art baseline global place recognition methods on the 4Seasons benchmark}. The gray line indicates the upper bound for the global place recognition accuracy.}
    \label{fig:gpr_results}
\end{figure}

As shown in Figure~\ref{fig:gpr_results}, we first plot two different recall curves: (1) \textit{Recall[\%] -- Threshold [m] @ Top1}: the recalls of different methods when changing the distance threshold in the range \SIrange[range-phrase = --]{1}{20}{\m} using only the top $1$ retrieved images, and (2) \textit{Recall[\%] -- Top N @ \SI{1}{\m}}: the recalls of different methods when changing the number of candidate retrievals $N \in \{1, 2, 3, \dots, 20\}$ using the fixed range bound \SI{1}{\m}. We also show the optimal recall by using the closest database images with respect to the ground-truth query image location as the candidates. This shows the upper bound of the global place recognition accuracy. 

From the results, we can see that NetVLAD still outperforms the other recent methods by a notable margin. One reason for NetVLAD's superior performance could be the introduction of the inductive bias into the network design, based on the established principle of classical VLAD~\citep{jegou2010aggregating,arandjelovic2013all}. However, one shall also admit that the gap between the state-of-the-art methods and the optimal performance is still quite large, and more research on global place recognition still needs to be conducted. 

We show the localization accuracy of the global place recognition methods without using local pose refinement. Note that for these methods we use the top $20$ candidates and loosened range bounds, namely, (\SI{1}{\m}, \SI{5}{\degree}) for \textbf{high precision}, (\SI{5}{\m}, \SI{10}{\degree}) for \textbf{medium precision}, and (\SI{10}{\m}, \SI{20}{\degree}) for \textbf{coarse precision}. The last three rows of Table~\ref{tab:localization_benchmark_results} show the individual global place recognition (GPR) performance on each of the evaluated scenarios from the 4Seasons benchmark.

\begin{figure}[t]
  \centering
  \includegraphics[width=0.8\linewidth]{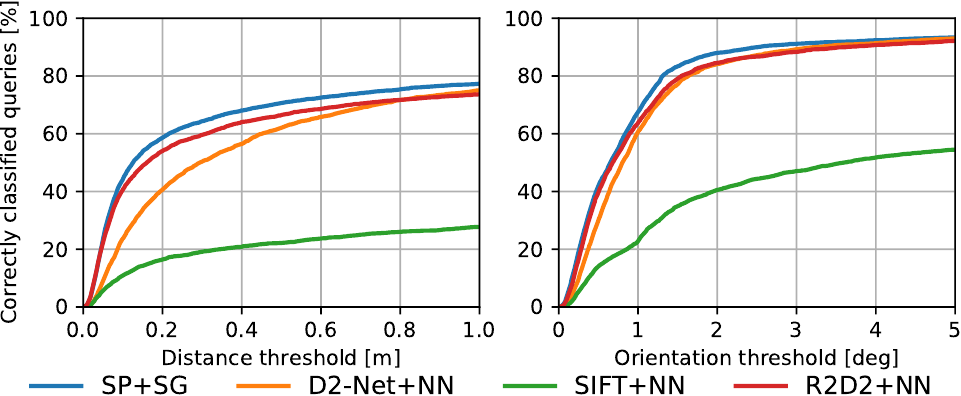}
  \caption{\textbf{Performance of state-of-the-art baseline map-based visual localization approaches on the 4Seasons benchmark.} The figure shows the cumulative localization accuracy against the translational, and rotational error, respectively.}
  \label{fig:reloc_tracking_results}
\end{figure}

\begin{table*}[t]
\begin{center}
\ra{1.3}
\caption{\textbf{Visual localization results on the 4Seasons benchmark.} We report the percentage of images localized within \SI{0.1}{\m} and \SI{1}{\degree}, \SI{0.25}{\m} and \SI{2}{\degree}, \SI{1}{\m} and \SI{5}{\degree} of the reference poses for map-based visual localization (MBVL) pipelines. For global place recognition (GPR) methods, we report the percentage of images localized within \SI{1}{\m} and \SI{5}{\degree}, \SI{5}{\m} and \SI{10}{\degree}, \SI{10}{\m} and \SI{20}{\degree} of the reference poses. The best-performing results for both MBVL and GPR pipelines are in bold.}
\label{tab:localization_benchmark_results}
\resizebox{\textwidth}{!}{
\begin{tabular}{@{}llccccccc|c@{}}
\toprule
\multicolumn{2}{c}{\textbf{Method}} & \bfseries Office Loop & \bfseries Neighborhood & \bfseries Business Campus & \bfseries Countryside & \bfseries City Loop & \bfseries Old Town & \bfseries Parking Garage & \bfseries Average\\
\midrule
\multirow{4}{*}{\begin{sideways}MBVL\end{sideways}}
& hloc~\citep{sarlin2019coarse} (SuperPoint~\citep{detone2018superpoint} + SuperGlue~\citep{sarlin2020superglue}) & 68.6 / 85.1 / 89.2 & 56.0 / 73.3 / 86.8 & 38.2 / 74.5 / 89.0 & 8.0 / 29.2 / 64.7 & 37.0 / 72.6 / 83.2 & 27.1 / 43.2 / 58.4 & 46.9 / 63.7 / 76.1 & \textbf{40.3} / \textbf{63.1} / \textbf{78.2}\\
& hloc~\citep{sarlin2019coarse} (D2-Net~\citep{Dusmanu2019CVPR} + NN) & 43.3 / 71.5 / 88.3 & 28.7 / 54.6 / 85.1 & 22.4 / 62.2 / 85.5 & 3.4 / 18.6 / 63.4 & 14.9 / 50.0 / 86.2 & 9.7 / 26.3 / 47.9 & 28.3 / 52.2 / 76.1 & 21.5 / 47.9 / 76.1\\
& hloc~\citep{sarlin2019coarse} (SIFT~\citep{lowe2004distinctive} + NN) & 24.6 / 39.3 / 52.6 & 33.6 / 50.6 / 66.1 & 3.5 / 9.9 / 18.9 & 0.1 / 0.6 / 2.4 & 9.0 / 22.3 / 40.3 & 0.0 / 0.3 / 4.0 & 9.7 / 16.8 / 27.4 & 11.5 / 20.0 / 30.2\\
& hloc~\citep{sarlin2019coarse} (R2D2~\citep{r2d2} + NN) & 66.4 / 83.1 / 88.3 & 59.8 / 81.9 / 96.0 & 36.9 / 68.6 / 82.9 & 3.8 / 18.1 / 53.6 & 36.1 / 77.2 / 92.0 & 16.5 / 27.8 / 40.0 & 41.6 / 66.4 / 78.8 & 37.3 / 60.4 / 75.9\\
\midrule
\multirow{3}{*}{\begin{sideways}GPR\end{sideways}}
& NetVLAD~\citep{arandjelovic2016netvlad} & 55.4 / 90.2 / 93.5 & 49.5 / 80.6 / 83.5 & -- & 10.6 / 29.9 / 34.1 & 24.6 / 54.0 / 62.3 & 30.8 / 64.3 / 79.2 & 37.9 / 79.3 / 86.2 & \textbf{34.8} / \textbf{66.4} / \textbf{73.1}\\
& CNN Image Retrieval (CIR)~\citep{radenovic2018fine} & 33.7 / 66.3 / 71.7 & 43.7 / 71.8 / 78.6 & -- & 6.8 / 22.0 / 28.8 & 6.7 / 25.0 / 31.3 & 14.9 / 45.7 / 62.0 & 27.6 / 55.2 / 65.5 & 22.2 / 47.7 / 56.3\\
& Deep Image Retrieval (DIR)~\citep{RARS19} & 26.1 / 58.7 / 68.5 & 41.7 / 72.8 / 79.6 & -- & 10.2 / 23.9 / 29.5 & 7.9 / 24.2 / 30.6 & 19.0 / 43.0 / 63.3 & 27.6 / 79.3 / 89.7 & 22.1 / 50.3 / 60.2\\
\bottomrule
\end{tabular}
}
\end{center}
\label{tab:vis_loc}
\end{table*}

\subsection{Map-Based Visual Localization}

For the evaluation of map-based visual localization, we use the following processing pipeline: we first build a \ac{sfm} model from the reference scene that provides correspondences between local features and 3D points in the reconstructed map. This is followed by 2D-3D matching between the query images and the database images. As the last step, those 2D-3D matches are used for camera pose estimation via \ac{pnp} and \ac{ransac}~\citep{fischler1981random}. In particular, we evaluate the current state-of-the-art coarse-to-fine hierarchical localization method~\citep{sarlin2019coarse} based on the following learned deep local feature descriptors: SuperPoint~\citep{detone2018superpoint}, D2-Net~\citep{Dusmanu2019CVPR}, and R2D2~\citep{r2d2}. Additionally, we use the classic \ac{sift}~\citep{lowe2004distinctive} algorithm. Hloc~\citep{sarlin2019coarse} simultaneously predicts local features and global descriptors for accurate \ac{6dof} localization. This approach first performs global place recognition to obtain location candidates, and afterward matches the local features only within those places. The extracted local image features are used to establish 2D-3D matches within a pre-built \ac{sfm} model. Pose estimation is performed using COLMAP~\citep{schonberger2016structure}. Therefore, this pipeline can be seen as a pose refinement strategy. For each scenario of the 4Seasons benchmark, we use a predefined recording as the reference map and a predefined recording as the query.

Figure~\ref{fig:reloc_tracking_results} shows the percentage of correctly classified queries when changing the distance and orientation thresholds, respectively. This figure shows the average performance for the different state-of-the-art map-based visual localization (MBVL) approaches across all evaluated scenarios. The first four rows of Table~\ref{tab:localization_benchmark_results} show the individual hierarchical localization performance on each of the evaluated scenarios from the 4Seasons benchmark.

From the results, we can see that the classic \ac{sift}+\ac{nn} approach shows a bad performance estimating the \ac{6dof} pose between the reference and query candidates. The results also suggest that deep learning-based algorithms dramatically outperform the classical method. This is due to the challenging nature of the benchmark since it deals with drastic lighting and illumination changes, occlusions, and changing environmental/weather conditions. Those results provide valuable insights into the limitations and failure cases of the different methods. We observe that learned feature descriptors significantly outperform classic methods under challenging conditions contained in the 4Seasons benchmark, with SuperPoint+SuperGlue yielding the best results overall. Nevertheless, the results from Table~\ref{tab:localization_benchmark_results} show that the long-term localization problem is still far from being solved, especially for highly dynamic environments (\eg \emph{Old Town}) and scenes that exhibit very similar structure (\eg \emph{Countryside}). 

Our benchmark provides the basis to enable more research advances that are needed to close this performance gap.

\section{Conclusion}

Current benchmarks either focus mainly on evaluating the performance of \acl{slam} algorithms or visual localization in isolation. To close this gap, we introduce a benchmark that by providing a holistic way for jointly benchmarking long-term visual \ac{slam} and localization. 

In this paper, we have introduced a comprehensive benchmark suite for visual \ac{slam} and visual localization for autonomous driving under challenging conditions. The benchmark covers a huge variety of environmental conditions, along with short-term and long-term weather and illumination changes. Moreover, we have reviewed and evaluated the current state-of-the-art baseline approaches for visual \ac{slam} and visual localization. We have observed large performance gaps and see huge potential in future work to close those gaps.

The benchmark dataset, evaluation server, and leaderboard will be available upon acceptance via the benchmark's website,~\url{https://go.vision.in.tum.de/4seasons}.

\begin{table*}[t]
\begin{center}
\ra{1.3}
\caption{\textbf{Statistics of the 4Seasons dataset.} This table shows the different scenarios and recordings along with the weather condition, seasons, and time of the day from our benchmark. We provide a variety of scenarios and short-term to long-term changes. These recordings are all released with ground truth (\ac{gnss}/\ac{imu}, point clouds, and reference poses) and can therefore be used for training learning-based techniques.}
\label{tab:4seasons_dataset_with_GT}
\resizebox{\textwidth}{!}{
\begin{tabular}{lllllll}
\toprule
\bfseries Scenario & \bfseries Recording & \bfseries \makecell[l]{Weather \\ (cloudy, rainy, snowy, sunny)} & \bfseries \makecell[l]{Season \\ (winter, spring, summer, fall)} & \bfseries \makecell[l]{Daytime \\ (morning, afternoon, evening, night)} & \bfseries \makecell[l]{Map Accuracy \\ Horizontal RMSE \\ (\ac{gnss}-Ref. Pose)} & \bfseries \makecell[l]{Map Accuracy \\ \% of Accurate Poses}\\
\midrule
office\_loop\_1\_train & 2020-03-24{\_}17-36-22 & sunny & spring & afternoon & \SI{6.84}{\cm} & \SI{85.93}{\percent}\\
office\_loop\_2\_train & 2020-03-24{\_}17-45-31 & sunny & spring & afternoon & \SI{6.34}{\cm} & \SI{86.92}{\percent}\\
office\_loop\_3\_train & 2020-04-07{\_}10-20-32 & sunny & spring & morning & \SI{5.44}{\cm} & \SI{77.72}{\percent}\\
office\_loop\_4\_train & 2020-06-12{\_}10-10-57 & sunny & summer & morning & \SI{2.74}{\cm} & \SI{54.01}{\percent}\\
office\_loop\_5\_train & 2021-01-07{\_}12-04-03 & cloudy/snowy & winter & afternoon & \SI{3.79}{\cm} & \SI{96.00}{\percent}\\
office\_loop\_6\_train & 2021-02-25{\_}13-51-57 & sunny & winter & afternoon & \SI{2.90}{\cm} & \SI{91.45}{\percent}\\
\midrule
neighborhood\_1\_train & 2020-03-26{\_}13-32-55 & cloudy & spring & afternoon & \SI{4.13}{\cm} & \SI{56.71}{\percent}\\
neighborhood\_2\_train & 2020-10-07{\_}14-47-51 & cloudy & fall & afternoon & \SI{1.19}{\cm} & \SI{85.00}{\percent}\\
neighborhood\_3\_train & 2020-10-07{\_}14-53-52 & rainy & fall & afternoon & \SI{2.00}{\cm} & \SI{84.12}{\percent}\\
neighborhood\_4\_train & 2020-12-22{\_}11-54-24 & cloudy & winter & morning & \SI{3.47}{\cm} & \SI{87.92}{\percent}\\
neighborhood\_5\_train & 2021-02-25{\_}13-25-15 & sunny & winter & afternoon & \SI{2.45}{\cm} & \SI{86.23}{\percent}\\
neighborhood\_6\_train & 2021-05-10{\_}18-02-12 & cloudy & spring & evening & \SI{1.74}{\cm} & \SI{69.43}{\percent}\\
neighborhood\_7\_train & 2021-05-10{\_}18-32-32 & cloudy & spring & evening & \SI{1.44}{\cm} & \SI{85.45}{\percent}\\
\midrule
business\_campus\_1\_train & 2020-10-08{\_}09-30-57 & sunny & fall & morning & \SI{5.49}{\cm} & \SI{83.08}{\percent}\\
business\_campus\_2\_train & 2021-01-07{\_}13-12-23 & cloudy/snowy & winter & afternoon & \SI{1.77}{\cm} & \SI{99.13}{\percent}\\
business\_campus\_3\_train & 2021-02-25{\_}14-16-43 & sunny & winter & afternoon & \SI{7.33}{\cm} & \SI{66.86}{\percent}\\
\midrule
countryside\_1\_train & 2020-04-07{\_}11-33-45 & sunny & spring & morning & \SI{3.96}{\cm} & \SI{90.89}{\percent}\\
countryside\_2\_train & 2020-06-12{\_}11-26-43 & sunny & summer & morning & \SI{2.54}{\cm} & \SI{87.00}{\percent}\\
countryside\_3\_train & 2020-10-08{\_}09-57-28 & sunny & fall & morning & \SI{1.94}{\cm} & \SI{89.37}{\percent}\\
countryside\_4\_train & 2021-01-07{\_}13-30-07 & cloudy/snowy & winter & afternoon & \SI{5.42}{\cm} & \SI{92.02}{\percent}\\
\midrule
city\_loop\_1\_train & 2020-12-22{\_}11-33-15 & rainy & winter & morning & \SI{6.85}{\cm} & \SI{83.08}{\percent}\\
city\_loop\_2\_train & 2021-01-07{\_}14-36-17 & snowy/sunny & winter & afternoon & \SI{4.76}{\cm} & \SI{84.27}{\percent}\\
city\_loop\_3\_train & 2021-02-25{\_}11-09-49 & sunny & winter & morning & \SI{3.41}{\cm} & \SI{85.14}{\percent}\\
\midrule
old\_town\_1\_train & 2020-10-08{\_}11-53-41 & cloudy & fall & morning & \SI{2.90}{\cm} & \SI{93.76}{\percent}\\
old\_town\_2\_train & 2021-01-07{\_}10-49-45 & cloudy/snowy/sunny & winter & morning & \SI{1.80}{\cm} & \SI{93.16}{\percent}\\
old\_town\_3\_train & 2021-02-25{\_}12-34-08 & sunny & winter & afternoon & \SI{1.43}{\cm} & \SI{83.12}{\percent}\\
old\_town\_4\_train & 2021-05-10{\_}21-32-00 & cloudy & spring & night & \SI{13.45}{\cm} & \SI{95.81}{\percent}\\
\midrule
parking\_garage\_1\_train & 2020-12-22{\_}12-04-35 & cloudy & winter & afternoon & \SI{1.43}{\cm} & \SI{33.22}{\percent}\\
parking\_garage\_2\_train & 2021-02-25{\_}13-39-06 & sunny & winter & afternoon & \SI{2.52}{\cm} & \SI{40.54}{\percent}\\
parking\_garage\_3\_train & 2021-05-10{\_}19-15-19 & cloudy & spring & evening & \SI{3.41}{\cm} & \SI{34.15}{\percent}\\
\bottomrule
\end{tabular}
}
\end{center}
\end{table*}

\backmatter

\bmhead{Acknowledgments}

We express our appreciation to our colleagues at Artisense for their help with setting up the recording setup and sensor design.

\section*{Declarations}

\bmhead{Competing Interests}

The authors declare that they have no conflict of interest.

\bibliographystyle{sn-basic}
\bibliography{sn-article}


\end{document}